\documentclass{article}

\usepackage[main, final]{neurips_2026}

\usepackage[utf8]{inputenc} 
\usepackage[T1]{fontenc}    
\usepackage{hyperref}       
\usepackage{url}            
\usepackage{booktabs}       
\usepackage{amsfonts}       
\usepackage{nicefrac}       
\usepackage{microtype}      
\usepackage{xcolor}         
\usepackage[dvipsnames]{xcolor}
\usepackage{multicol}
\usepackage{multirow}
\usepackage{wrapfig}
\usepackage{subcaption}
\usepackage{amsmath}
\usepackage{pifont}                         
\usepackage{graphicx}
\usepackage{fancyhdr,amssymb}
\usepackage[ruled,vlined]{algorithm2e}
\usepackage{circledsteps}
\usepackage{paralist}

\definecolor{GainBright}{RGB}{0,114,178}   

\usepackage{colortbl}
\usepackage{makecell}
\usepackage{microtype}
\usepackage{placeins}

\usepackage{minitoc}

\usepackage{amssymb,amsmath,amsthm}
\usepackage{pgf}
\usepackage{pgfplots}

\usepackage{transparent}
\usepackage{adjustbox}

\usepackage{algorithmic}

\def\our{InTAct}
\definecolor{myPurple}{RGB}{140,26,245}
\definecolor{avgcolor}{HTML}{E0F2F1}
\newcommand{\res}[2]{#1{\pm}\scriptstyle #2}

\title{InTAct: Interval-based Task Activation Consolidation for Continual Learning}

\author{%
  \renewcommand{\And}{\end{tabular}\hspace{1.5em}\begin{tabular}[t]{c}}
  Patryk Krukowski\thanks{\texttt{patryk.krukowski@doctoral.uj.edu.pl}} \\
  \small Doctoral School of Exact and Natural Sciences \\
  \small Jagiellonian University \\
  \small Akces NCBR
  \And
  Jan Miksa \\
  \small Doctoral School of Exact and Natural Sciences \\
  \small Jagiellonian University \\
  \small IDEAS Research Institute
  \AND
  Piotr Helm \\
  \small Doctoral School of Exact and Natural Sciences \\
  \small Jagiellonian University
  \And
  Jacek Tabor \\
  \small Jagiellonian University \\
  \small Centre for Credible AI
  \AND
  Paweł Wawrzyński \\
  \small IDEAS Research Institute
  \And
  Przemysław Spurek \\
  \small Jagiellonian University \\
  \small IDEAS Research Institute
}

\begin{document}

\maketitle

\begin{abstract}
  Continual learning seeks to acquire new knowledge while preserving previously learned representations. Despite the success of prompt-based methods, they remain fragile in domain-incremental learning scenarios, where shifts in the input distribution cause representation drift in shared layers and lead to forgetting. We introduce \our{}\footnote{https://github.com/pkrukowski1/Prompt\_InTAct}, a method that preserves functional behavior in shared representations without freezing parameters or retaining past data. \our{} constrains updates within activation regions associated with prior domains, while allowing flexible adaptation elsewhere, thereby stabilizing neuron functionality rather than directly restricting parameter values. The approach is architecture-agnostic, integrates seamlessly with existing prompt-based frameworks, and matches or outperforms state-of-the-art methods on domain-incremental benchmarks.
\end{abstract}

\section{Introduction}
\label{sec:intro}

Continual learning aims to enable neural networks to acquire new knowledge while retaining previously learned information. This requires balancing stability, the retention of past knowledge, with plasticity, the ability to adapt to new tasks. Unlike humans, deep neural networks often struggle to maintain this balance~\cite{Kirkpatrick_2017}. As they adapt to new data distributions, their internal representations may drift, altering features important for earlier tasks and ultimately leading to catastrophic forgetting~\cite{MCCLOSKEY1989109}.

A large body of prior work addresses this issue by constraining parameter updates through regularization, architectural modifications, or optimization-based strategies~\cite{2019parisi+4,2021delange+7,2022masana+5}. While effective to some extent, these methods predominantly operate in the parameter space and rarely provide explicit control over how the network’s function evolves across tasks. Moreover, despite the large number of continual learning strategies, most existing approaches lack rigorous mathematical guarantees against forgetting. Methods that do provide formal assurances typically rely on \textit{interval arithmetic} (IA) in the weight space, as exemplified by frameworks such as InterContiNet~\cite{wolczyk2022continual} and HINT~\cite{2025krukowski+5}. However, constraining high-dimensional parameter updates is computationally expensive and does not necessarily ensure the model’s functional stability.

\begin{figure*}[t]
\centering
    \begin{tikzpicture}[scale=0.22]
    \node[inner sep=0pt] (russell) at (-5.0,0)
    {\includegraphics[trim={0,1cm, 0,5cm, 0,1cm, 0,8cm},clip,width=0.99\linewidth]{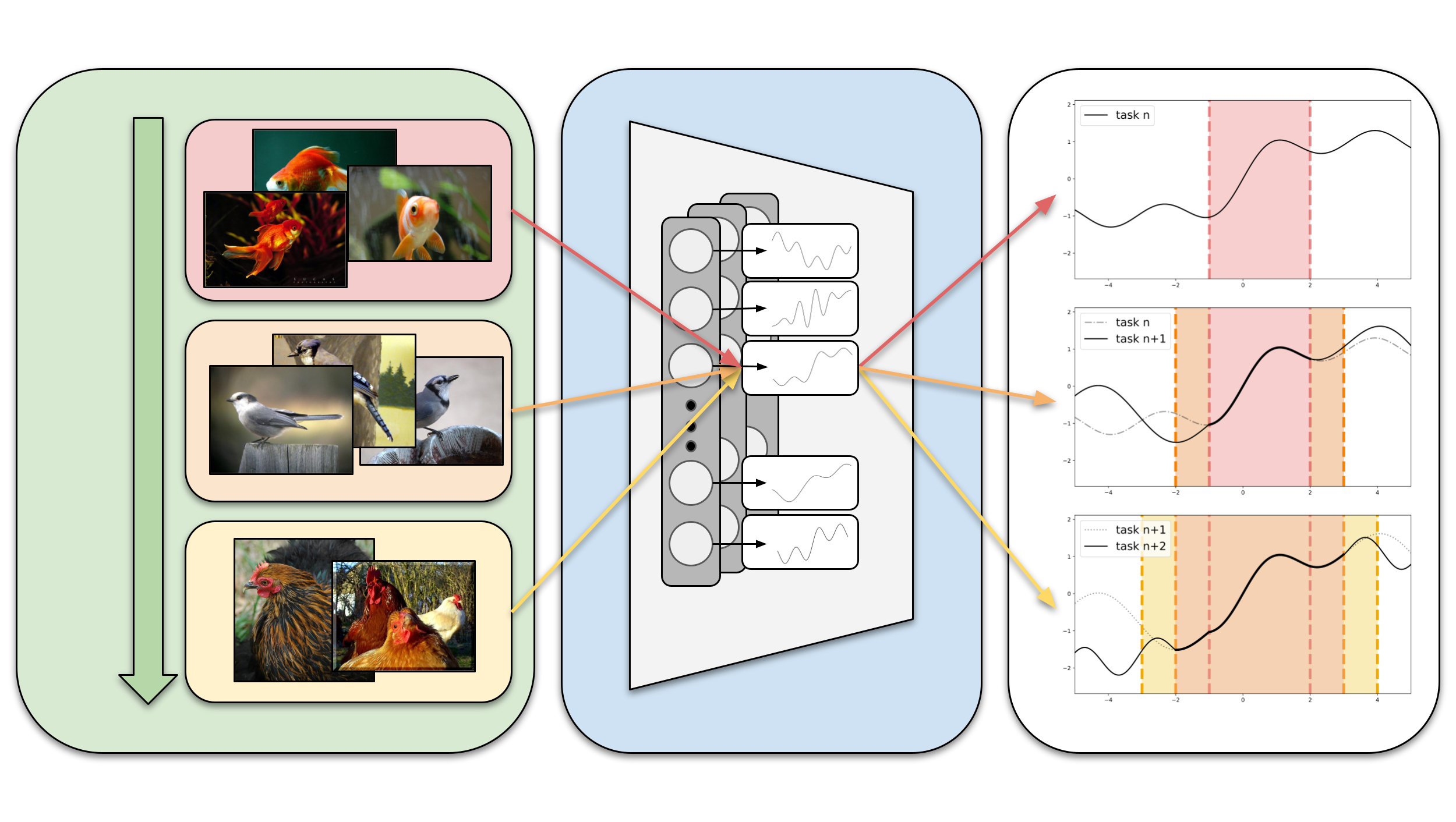} };

    \node[scale=0.9] at (-25, 16) {Tasks};    
    \node[scale=0.6] at (-33.5,9.5) {Task $N$};    
    \node[scale=0.6] at (-33.2,1) {Task $N{+}1$};    
    \node[scale=0.6] at (-33.2,-7.5) {Task $N{+}2$};  

    \node[scale=0.6] at (-17.8,5.6) {class: goldfish};    
    \node[scale=0.6] at (-17.5, -3) {class: bird};    
    \node[scale=0.6] at (-17.5,-11.6) {class: rooster};  

    \node[scale=0.9] at (-4, 16) {Model};    
    \node[scale=0.6] at (-6.9, 10) {Layers};   

    \node[scale=0.9] at (16, 16) {In-Layer Transformation};
    
    \end{tikzpicture}
    \caption{\textbf{Method Overview.} During training on task $(N{+}1)$, \our{} preserves the functional region established by task $N$ (pink) while allowing the model to learn new knowledge in a distinct region (orange). After training, these regions merge into an expanded protected region (yellow), defining where the layer’s transformation should remain stable in future updates. The dashed curves illustrate transformations from earlier tasks, highlighting that their behaviors remain preserved within their respective regions, while adaptation is freely allowed outside them.}
    \label{fig:teaser}
\end{figure*}

To address these limitations, we introduce \our{} (Interval-based Task Activation Consolidation), a novel approach that enforces functional stability by directly regulating neural activations rather than model parameters. \our{} identifies neuron-wise activation intervals that characterize stable operating regions established during previous tasks. These intervals are aggregated at the layer level to form multidimensional hypercubes, which delineate regions of the feature space corresponding to previously acquired knowledge (see Fig.~\ref{fig:teaser}). During subsequent training, \our{} constrains model updates within these regions while permitting unrestricted adaptation outside them. By enforcing prediction invariance whenever internal signals remain within the preserved activation intervals, \our{} provides a tractable and efficient layer-wise mathematical guarantee of functional stability, which avoids the excessive conservativeness of end-to-end interval propagation.

When learning new tasks, the resulting hypercubes act as elastic boundaries: they encourage the model to remain within trusted regions for familiar concepts, while still allowing exploration beyond these regions to accommodate novel information. This mechanism naturally balances stability and plasticity. Crucially, instead of restricting the model’s parameters, \our{} directly regulates activations, which constitute the model’s observable functional behavior. This results in a more direct and effective control over how the network evolves over time.

Although the proposed framework is architecture-agnostic and applicable to a wide range of continual learning scenarios, its integration with prompt-based continual learning methods~\cite{wang2022L2P,wang2022DualPrompt} proves particularly effective. In this setting, \our{} achieves state-of-the-art (SOTA) performance on challenging domain-incremental benchmarks. Our evaluations confirm that \our{} effectively stabilizes internal representations and consistently surpasses baselines without the need for data replay. The main contributions of this work are:
\begin{enumerate}
    \item A novel continual learning framework, \our{}, which enforces functional stability by constraining neuron activations through interval-based representations rather than parameter-level restrictions.
    \item We introduce an IA-based formulation together with a formal layer-wise stability theorem showing that \our{} preserves functional behavior locally within protected regions of the feature space while maintaining plasticity outside these regions.
    \item \our{} seamlessly integrates with prompt-based architectures, achieving SOTA performance on domain-incremental benchmarks without replay or parameter regularization.\footnotemark
\end{enumerate}
\footnotetext{Integration details are provided in Appendix~\ref{appendix:sec_intact_in_other_methods}.}

\section{Related Works}\label{sec:related_works}
For an extended overview, see Appendix~\ref{appendix:sec_related_work}.
\paragraph{Continual Learning.}
Traditional continual learning methods follow three main paradigms. \textit{Replay-based} approaches~\cite{2018chaudhry+3,2020buzzega+4,2025urettini+1} retain exemplars or generate synthetic data~\cite{2017shin+3} to approximate past distributions, at the cost of additional memory. \textit{Regularization-based} methods~\cite{Kirkpatrick_2017,li2017learningforgetting,2022wang+4} preserve previous knowledge by constraining parameter updates through penalties or projections, but often struggle to maintain stability over long task sequences. Finally, \textit{parameter-isolation} approaches~\cite{2018mallya+1,2018serra+3,liang2024InfLoRA} allocate task-specific parameters or subspaces, often relying on task IDs at inference time.

\paragraph{Prompt-Based Continual Learning.}
Recent architectures leverage frozen pre-trained backbones, directing adaptation via learnable prompts. L2P~\cite{wang2022L2P} and DualPrompt~\cite{wang2022DualPrompt} utilize a pool of keys and prompts to query task-specific knowledge. CODA-Prompt~\cite{smith2023CODAPrompt} extends this paradigm via attention-based decomposition, while C-Prompt~\cite{Liu2024C_Prompt} shares compositional prompt components across domains. Despite their success, these methods remain vulnerable to \textit{representational drift} in shared components, particularly prompt keys and the classification head. This drift is especially detrimental in domain-incremental learning (DIL), where task boundaries are inherently ambiguous. KA-Prompt~\cite{xu2025comp} observes that prompts learned for different domains are misaligned component-wise, so fusing them across domains mixes conflicting knowledge. To address this, it initializes new prompts from relevant historical ones identified via greedy search and encourages consistency with them during training.

\paragraph{Interval Arithmetic in Continual Learning.}
While IA~\cite{dahlquist2008numerical,moore2009introduction} offers a principled way to obtain rigorous numerical bounds, its use in continual learning remains largely unexplored. Existing methods primarily apply IA in the \textit{parameter space}. InterContiNet~\cite{wolczyk2022continual} identifies safe weight intervals for multi-task compatibility, while HINT~\cite{2025krukowski+5} uses hypernetworks to map interval-based embeddings to network parameters. However, propagating intervals through deep networks suffers from the \textit{wrapping effect}~\cite{moore2009introduction}, which causes bounds to expand with network depth and become excessively loose. Rather than performing interval propagation, \our{} uses IA locally to construct static \textit{hypercubes} around representations of previously learned tasks. These bounds are then used to regularize subsequent representations, promoting functional stability without the computational cost and bound explosion of full interval propagation.

\begin{figure*}[t]
  \centering
  \begin{subfigure}[b]{0.32\linewidth}
    \centering
    \includegraphics[width=\linewidth]{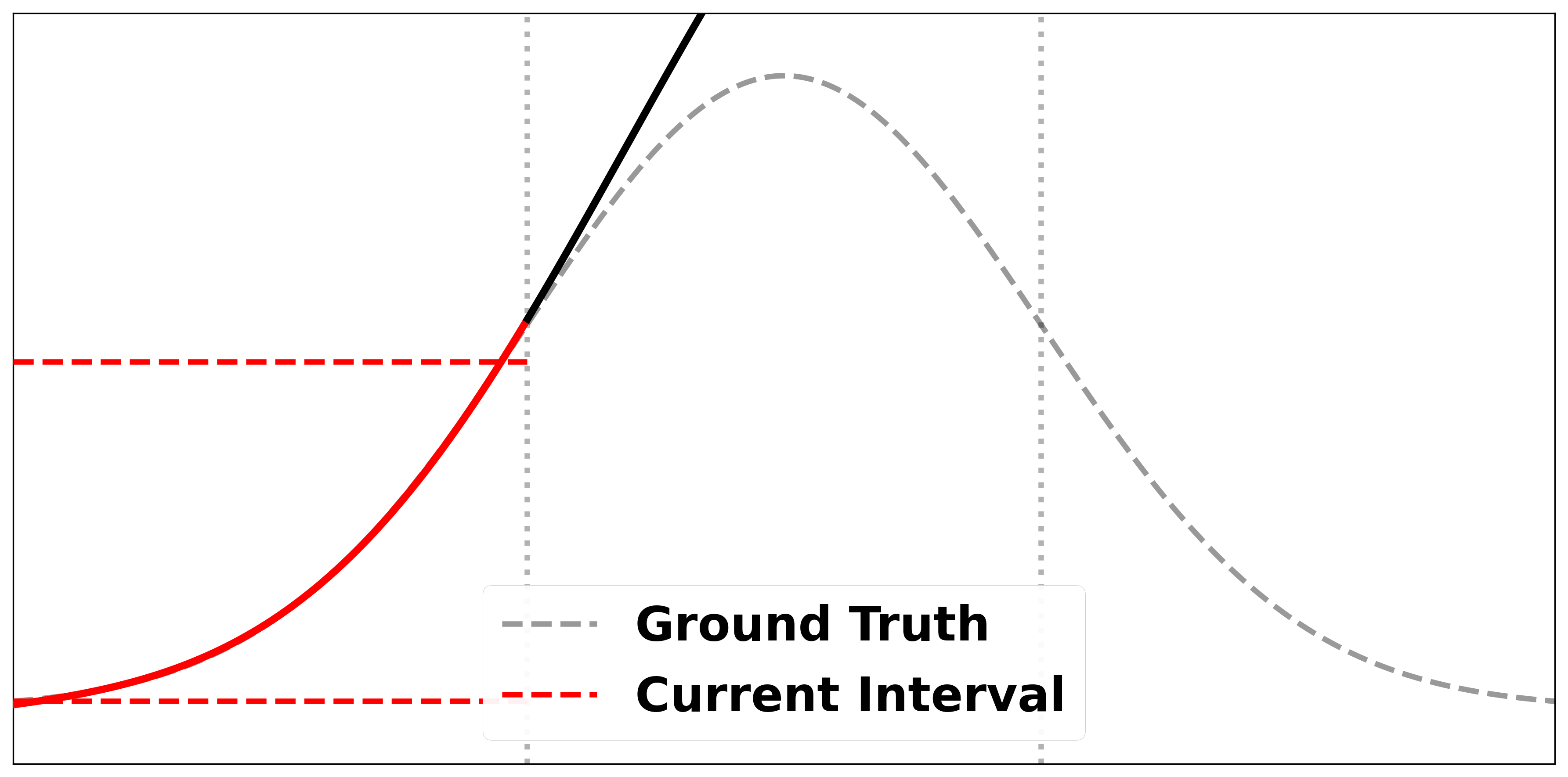}
    \caption{Task 1: Initial learning and interval computation.}
    \label{fig:gaussian_task_1}
  \end{subfigure}
  \hfill
  \begin{subfigure}[b]{0.32\linewidth}
    \centering
    \includegraphics[width=\linewidth]{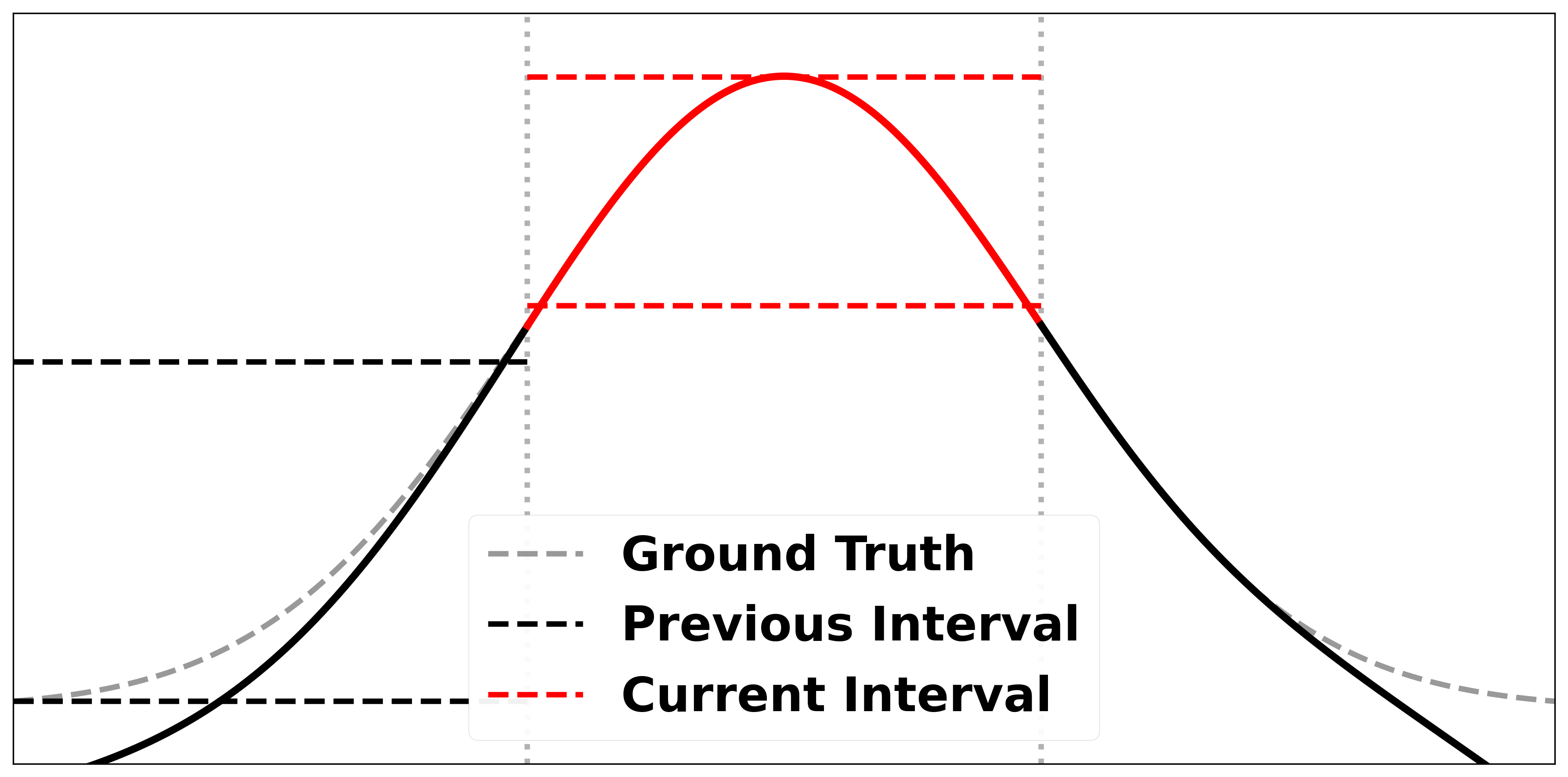}
    \caption{Task 2: Adaptation with stability constraints.}
    \label{fig:gaussian_task_2}
  \end{subfigure}
  \hfill
  \begin{subfigure}[b]{0.32\linewidth}
    \centering
    \includegraphics[width=\linewidth]{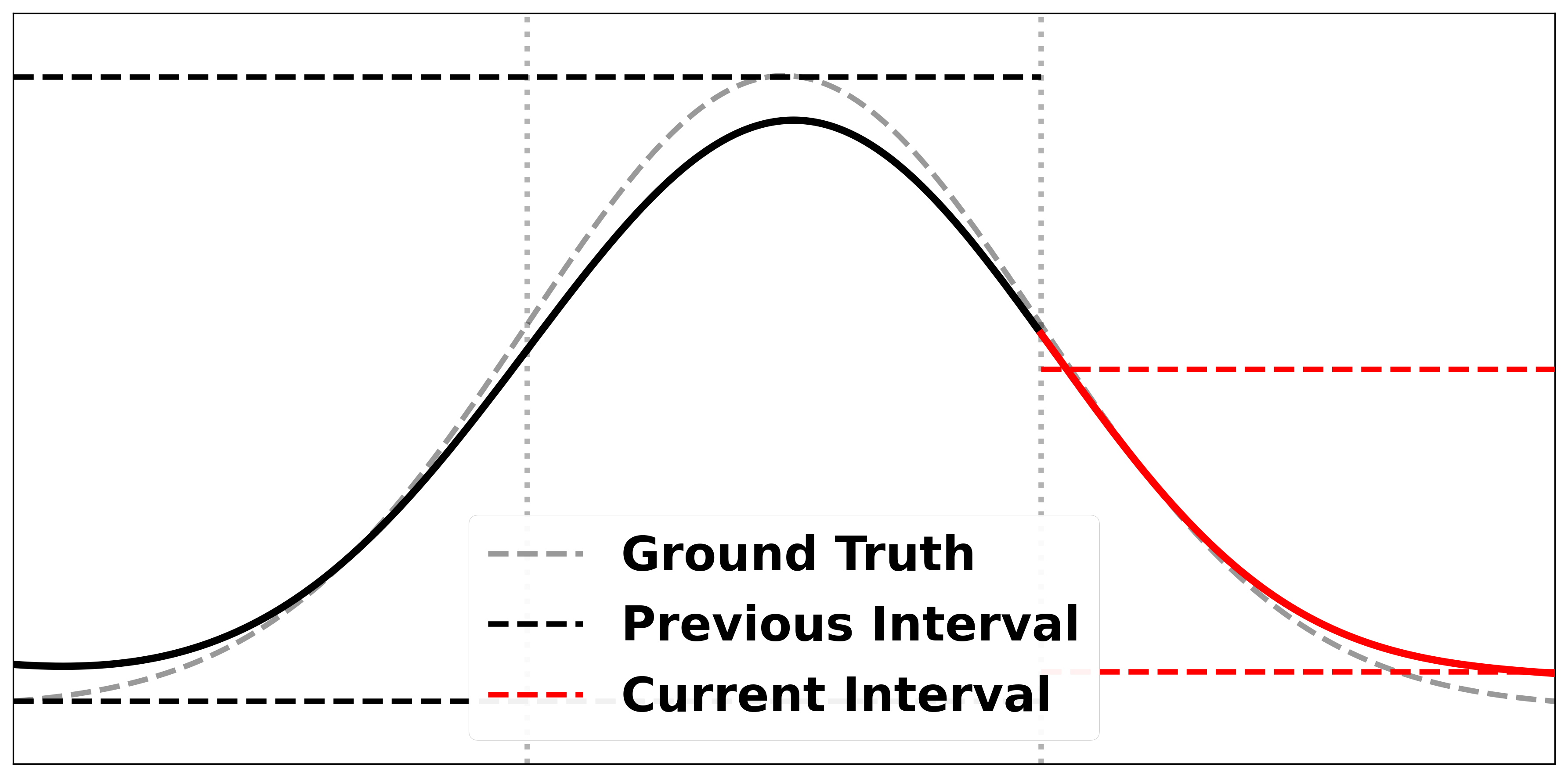}
    \caption{Task 3: Global accumulation without forgetting.}
    \label{fig:gaussian_task_3}
  \end{subfigure}
  \vspace{-2mm}
  \caption{
    \textbf{Illustration of localized functional preservation in \our{}.}
    A model is trained sequentially on three tasks to approximate different regions of a Gaussian-shaped function. The dashed lines denote the protected intervals (1D hypercubes) estimated after each task. During subsequent training stages, these intervals constrain functional drift within previously occupied regions while allowing the function to adapt freely in unexplored regions of the input space. As a result, \our{} preserves prior knowledge locally without restricting plasticity globally.
    }
  \label{fig:gaussian}
  \vspace{-3mm}
\end{figure*}

\section{Method}
\label{sec:method}

\our{} mitigates forgetting by protecting geometric regions of the feature space instead of regularizing model weights. Specifically, it constructs deterministic bounding boxes around historical representations, guaranteeing invariant classifier outputs within protected regions while preserving representation consistency across tasks.

\paragraph{Conceptual Overview.}
Fig.~\ref{fig:gaussian} illustrates the intuition behind \our{}. After learning Task~1, the method constructs a protected region around its representations. During training on Task~2, \our{} enforces functional invariance within this region while allowing unconstrained adaptation outside it. As a result, the model can acquire new knowledge in previously unused regions of the feature space without disrupting prior tasks, naturally balancing stability and plasticity.

Our framework is designed for prompt-based continual learning, where learnable prompts adapt a frozen backbone to produce task-specific representations. Although the backbone itself remains fixed, both the prompt pool and the shared classifier evolve continuously across tasks, making the model susceptible to representation drift and catastrophic forgetting. \our{} mitigates this effect by directly constraining the geometry of the intermediate feature space. Since final predictions depend entirely on these representations, the proposed framework jointly preserves classifier behavior within historical feature regions while anchoring prompts to ensure that previously learned samples remain mapped to these protected regions. This design preserves prior knowledge while maintaining the flexibility required to learn new tasks, as confirmed by our experimental results. Importantly, the proposed activation-space constraints are architecture-agnostic and can be easily integrated into existing continual learning frameworks (see Appendix~\ref{appendix:sec_intact_in_other_methods}).

\subsection{Activation Interval Representation}
\label{sec:activation_interval}

To preserve historical knowledge without storing replay samples, we construct a compact geometric representation of the feature distribution. This representation is applied both to the \texttt{[CLS]} token embeddings and to intermediate classifier activations.

Let $\mathcal{Z}_{t} \subset \mathbb{R}^{d}$ denote the set of feature representations extracted from task $t$. Rather than modeling the full distribution explicitly, we approximate its support using a quantile-based hypercube that captures the dominant activation region while reducing the influence of extreme values. Specifically, we retain the central part of the distribution and exclude the tails beyond the $(1-\alpha)$- and $\alpha$-quantiles, where $\alpha \in (0.5, 1]$. Unless stated otherwise, we set $\alpha = 0.95$.

For each feature dimension $j$, we define the task-specific activation interval $[\underline{z}_{t,j}, \overline{z}_{t,j}]$. The corresponding task-specific hypercube is obtained as the Cartesian product of these intervals:
\begin{equation*}
    \mathcal{H}_{t}
    =
    \big[
        \underline{\mathbf{z}}_{t},
        \overline{\mathbf{z}}_{t}
    \big]
    =
    \prod_{j=1}^{d}
    \big[
        \underline{z}_{t,j},
        \overline{z}_{t,j}
    \big],
\end{equation*}
where $\underline{z}_{t,j}$ and $\overline{z}_{t,j}$ are the $(1-\alpha)$- and $\alpha$-quantiles, respectively, of the activations $\{z_j : \mathbf{z} \in \mathcal{Z}_{t}\}$ along dimension $j$. Throughout the paper, we use the term \textit{hypercube} to refer to this axis-aligned box, which may have different side lengths across dimensions. Accordingly, $\mathbf{z} \in [\underline{\mathbf{z}}, \overline{\mathbf{z}}]$ denotes that $\underline{z}_j \leq z_j \leq \overline{z}_j$ for every $j$.

To preserve constraints from previously learned tasks without retaining past data, we maintain a cumulative hypercube $\mathcal{H}^{(\leq t)}$ that aggregates all historical intervals. After completing task $t$, the global boundary is updated as
\begin{equation*}
    \mathbf{\mathcal{H}}^{(\leq t)}
    =
    \Big[
        \min\big(
            \mathbf{\underline{z}}^{(\leq t-1)},
            \mathbf{\underline{z}}_{t}
        \big),
        \max\big(
            \mathbf{\overline{z}}^{(\leq t-1)},
            \mathbf{\overline{z}}_{t}
        \big)
    \Big],
\end{equation*}
where $[\underline{\mathbf{z}}^{(\leq t-1)}, \overline{\mathbf{z}}^{(\leq t-1)}] = \mathcal{H}^{(\leq t-1)}$, and $\min(\cdot,\cdot)$ and $\max(\cdot,\cdot)$ denote element-wise minimum and maximum operators, respectively. This update computes the interval hull $\mathcal{H}^{(\leq t)} = \mathcal{H}^{(\leq t-1)} \sqcup \mathcal{H}_{t}$, i.e., the smallest hypercube enclosing the task-specific hypercubes of all previously observed tasks. At the same time, the resulting geometric representation remains memory-efficient, as storing the hypercube requires only lower and upper bounds for each feature dimension. Consequently, the memory footprint grows linearly with the feature dimensionality while remaining independent of the number of stored samples or tasks.

\subsection{Functional Preservation via Regularization}
\label{sec:regularization}

We now describe how \our{} preserves classifier behavior within protected regions of the feature space. For clarity, we first consider a single linear layer. In deep architectures, the same mechanism is applied at multiple stages of the network by constructing protected hypercubes directly in the intermediate activation spaces. Consequently, we address stability in the full model by controlling the functional drift induced by each of these linear transformations separately. For notational simplicity, we omit explicit layer indices in the following derivation.

Consider a layer $f(\mathbf{z}) = \mathbf{W}\mathbf{z} + \mathbf{b}$,
where $\mathbf{z}$ denotes an input representation, such as a \texttt{[CLS]} token or an intermediate hidden activation. During continual learning, the layer parameters evolve from $(\mathbf{W},\mathbf{b})$ to $(\mathbf{W}',\mathbf{b}')$ as new tasks are acquired. The resulting functional change induced at representation $\mathbf{z}$ can therefore be written as
\[
\Delta f(\mathbf{z})
=
\Delta \mathbf{W}\mathbf{z} + \Delta \mathbf{b},
\]
where $\Delta \mathbf{W} = \mathbf{W}' - \mathbf{W}$, $\Delta \mathbf{b} = \mathbf{b}' - \mathbf{b}$, and $\Delta f_i(\mathbf{z})$ denotes the $i$-th component of $\Delta f(\mathbf{z})$.

Our objective is to bound this drift within regions associated with previously learned representations, thereby preserving historical decision behavior while allowing adaptation outside the protected regions. Extensions to other layer types, along with a discussion of why we do not apply the regularizer to nonlinear functions, are provided in Appendix~\ref{appendix:sec_internal_representation_drift_loss_for_other_layers}.

\paragraph{Tractability via Interval Arithmetic.}

Directly enforcing stability for every point inside the protected region $\mathbf{\mathcal{H}}^{(\leq t-1)}$ is computationally intractable. Instead, we employ \emph{interval arithmetic (IA)} to derive rigorous bounds on the maximum functional drift over the entire region simultaneously.

Let $\mathcal{H}^{(\leq t-1)} = [\underline{\mathbf{z}}, \overline{\mathbf{z}}]$ denote the cumulative activation hypercube up to task $t-1$, where we omit the superscripts on its bounds for brevity. For each output unit $i$, we decompose the corresponding weight update $\Delta \mathbf{w}_i$ (the $i$-th row of $\Delta \mathbf{W}$) into its positive and negative components:
\[
(\Delta \mathbf{w}_i)^+ = \max(\mathbf{0}, \Delta \mathbf{w}_i),
\qquad
(\Delta \mathbf{w}_i)^- = \max(\mathbf{0}, -\Delta \mathbf{w}_i),
\]
where the maximum is taken element-wise. Using interval propagation (see Appendix~\ref{appendix:derivation_internal_representation_drift}), the exact lower and upper bounds of the induced output drift over $\mathcal{H}^{(\leq t-1)}$ are given by
\begin{align*}
    \underline{\delta}_{i}
    &=
    \min_{\mathbf{z} \in \mathcal{H}^{(\leq t-1)}} \Delta f_i(\mathbf{z})
    =
    (\Delta \mathbf{w}_i)^+ \underline{\mathbf{z}}
    -
    (\Delta \mathbf{w}_i)^- \overline{\mathbf{z}}
    +
    \Delta b_i, \\
    \overline{\delta}_{i}
    &=
    \max_{\mathbf{z} \in \mathcal{H}^{(\leq t-1)}} \Delta f_i(\mathbf{z})
    =
    (\Delta \mathbf{w}_i)^+ \overline{\mathbf{z}}
    -
    (\Delta \mathbf{w}_i)^- \underline{\mathbf{z}}
    +
    \Delta b_i.
\end{align*}

We define the internal representation drift loss ($\mathcal{L}_{\mathrm{IntDrift}}$) as
\begin{equation}
\label{eq:reg_int_repr}
\mathcal{L}_{\mathrm{IntDrift}}
=
\frac{\lambda_{\mathrm{IntDrift}}}{N}
\sum_{i=1}^{N}
\Big(
    (\underline{\delta}_{i})^2
    +
    (\overline{\delta}_{i})^2
\Big),
\end{equation}
where $N$ is the number of output units in the considered layer and $\lambda_{\mathrm{IntDrift}}$ controls the regularization strength. Minimizing this loss penalizes deviations of the output bounds across all units, thereby limiting worst-case functional drift within the protected activation region. For brevity, we omit the explicit arguments of the loss functions throughout the paper.

\paragraph{Mathematical Stability Guarantee.}

Classical regularization-based continual learning methods constrain parameter updates in order to preserve previous knowledge. In contrast, \our{} directly constrains the functional behavior of the model inside regions occupied by historical representations. This is achieved by penalizing deviations in the model's output bounds through $\mathcal{L}_{\mathrm{IntDrift}}$.

\newtheorem{theorem}{Theorem}
\begin{theorem}
\label{thm:math_guarantee}
Let $\mathcal{H} = [\underline{\mathbf{z}}, \overline{\mathbf{z}}]$ be the protected hypercube with strictly positive dimensional diameters $\mathrm{diam}_j(\mathcal{H}) = \overline{z}_j - \underline{z}_j$. If the internal drift loss satisfies $\mathcal{L}_{\mathrm{IntDrift}} \le \varepsilon$ for some $\varepsilon > 0$, then the functional drift for any representation $\mathbf{z}$ and output unit $i$ is bounded by
\[
|\Delta f_i(\mathbf{z})|
\leq
\mathcal{O}(\sqrt{\varepsilon})
+
\sum_{j=1}^{d}
\left(
\frac{d_j(\mathbf{z})}
{\mathrm{diam}_j(\mathcal{H})}
\right)
\mathcal{O}(\sqrt{\varepsilon}),
\]
where
\[
d_j(\mathbf{z})
=
\min_{\mathbf{z}'\in\mathcal{H}}
|z_j - z'_j|
\]
denotes the distance from $\mathbf{z}$ to the historical hypercube along the $j$-th dimension. The implicit multiplicative constants in $\mathcal{O}(\cdot)$ depend on $N$ and $\lambda_{\mathrm{IntDrift}}$ (see Appendix~\ref{appendix:sec_proofs}).
\end{theorem}

Theorem~\ref{thm:math_guarantee} establishes a localized, layer-wise upper bound on functional drift in the feature space. For representations inside the protected region ($d_j(\mathbf{z}) = 0$ for all $j$), the bound reduces to $\mathcal{O}(\sqrt{\varepsilon})$. Outside this region, the bound increases with the normalized distance from the historical hypercube, allowing greater functional flexibility farther from previously observed regions. Since the bound holds for any representation $\mathbf{z}$, it also covers historical activations excluded by the quantile-based construction. Empirically, most activation coordinates of previously learned tasks remain within the cumulative hypercube throughout training, with coverage ranging from $96.1\%$ to $97.0\%$ after the final task on ImageNet-R (Tab.~\ref{tab:hypercube_coverage} in Appendix~\ref{appendix:additional_results}).

In our implementation, interval regularization is applied to the affine layers of the classifier. Extending interval propagation through the entire ViT backbone would suffer from the wrapping effect, resulting in excessively conservative bounds~\cite{gowal2019effectivenessintervalboundpropagation}. Consequently, the theorem guarantees localized functional stability in the classifier rather than providing an end-to-end guarantee for the entire model. Preserving the backbone representations is addressed through complementary regularization mechanisms introduced next. Together, these mechanisms promote stability while retaining the flexibility needed to learn new tasks. The full proof is provided in Appendix~\ref{appendix:sec_proofs}.

\paragraph{Feature Space Regularization.}
For the guarantee in Theorem~\ref{thm:math_guarantee} to remain effective in prompt-based architectures, historical representations must remain within their protected regions throughout continual learning. However, continuously updating prompts can cause representations of previously learned samples to drift, potentially moving them outside the protected hypercube and weakening the resulting stability guarantee.

To mitigate this effect, we introduce a region-selective feature consistency objective that anchors historical representations within their protected regions. Specifically, we define the coordinate-wise binary gate $m_{i,j} = \mathbb{I}[\underline{z}_{j} \leq z_{i,j}^{(t)} \leq \overline{z}_{j}]$, where $\mathbb{I}[\cdot]$ denotes the indicator function and $[\underline{\mathbf{z}}, \overline{\mathbf{z}}] = \mathcal{H}^{(\leq t-1)}$. This gate enforces consistency only on the coordinates of the current representation that lie within the protected hypercube (a per-sample gate would rarely be active; see Appendix~\ref{appendix:additional_results}). Additional analysis of how the clipping threshold $\alpha$ affects the stability--plasticity trade-off is provided in Appendix~\ref{appendix:additional_results} (Tab.~\ref{appendix:tab_alpha_sensitivity}). We then define the feature regularization term as
\begin{equation}
\label{eq:prompt_reg_loss}
\mathcal{L}_{\mathrm{Feat}}
=
\frac{\lambda_{\mathrm{Feat}}}{\sum_{i,j} m_{i,j}}
\sum_{i=1}^{N_B}
\sum_{j=1}^{d}
m_{i,j}
\big(
z_{i,j}^{(t)}
-
z_{i,j}^{(t-1)}
\big)^2,
\end{equation}
where $N_B$ denotes the batch size, $\mathbf{z}_i^{(t-1)}$ and $\mathbf{z}_i^{(t)}$ are the representations of the $i$-th sample in the current training batch, produced by the frozen model after task $t-1$ and the current model, respectively. No data from previous tasks is replayed, and $\lambda_{\mathrm{Feat}}$ controls the regularization strength. Consequently, the objective preserves the coordinates within the protected region while leaving those outside it free to adapt to new tasks.

The feature consistency objective complements Theorem~\ref{thm:math_guarantee}. For an input with representations $\mathbf{z}^{(t-1)}$ and $\mathbf{z}^{(t)}$, the output change of a protected layer is ${f'(\mathbf{z}^{(t)}) - f(\mathbf{z}^{(t-1)})} = {\Delta f(\mathbf{z}^{(t)})} + {\mathbf{W}(\mathbf{z}^{(t)} - \mathbf{z}^{(t-1)})}$. The theorem bounds the first term, and coordinates inside $\mathcal{H}^{(\leq t-1)}$ do not contribute to its distance-dependent part, while $\mathcal{L}_{\mathrm{Feat}}$ limits the contribution of exactly these coordinates to the second term. Thus, minimizing $\mathcal{L}_{\mathrm{Feat}}$ encourages the current model to preserve the frozen model's representations inside the protected region, without explicitly storing or replaying historical samples.

\begin{figure*}[t]
    \centering
    \begin{tikzpicture}[scale=0.22]
    \node[inner sep=0pt] (russell) at (-18.0,0)
    {\includegraphics[trim={0cm, 18cm, 0cm, 0cm},clip,width=1.0\linewidth]{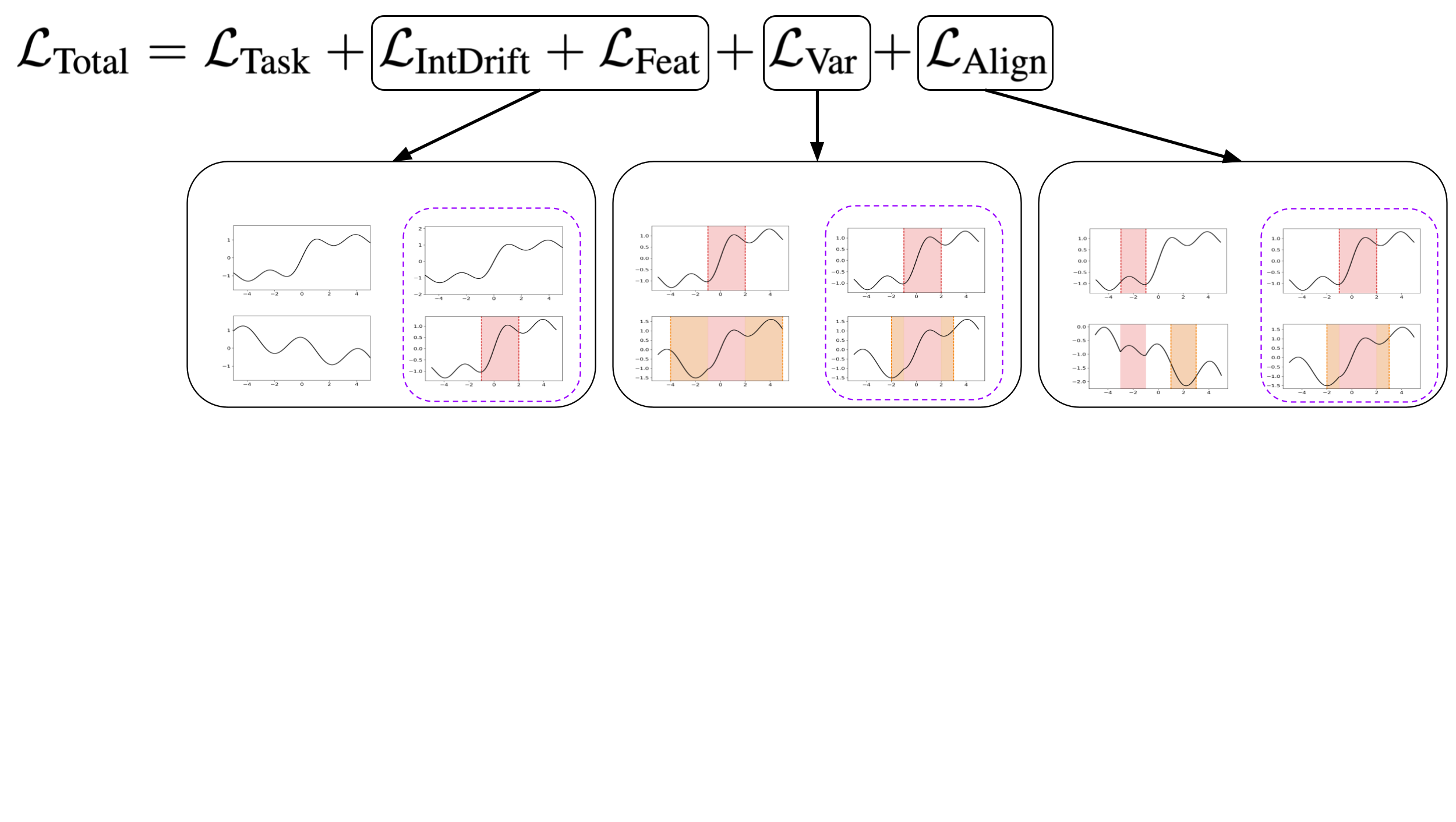} };

    \node[text=myPurple] at (-28.3, 3.3) {\textbf{\our{}}};
    \node[text=myPurple] at (-9.8, 3.3) {\textbf{\our{}}};
    \node[text=myPurple] at (9.3, 3.3) {\textbf{\our{}}};

    \node at (-46, 0) {Task $N$};
    \draw[dotted] (-50, -2) -- (14,-2);
    \node at (-46, -4.2) {Task $N+1$};

    \end{tikzpicture}
    \vspace{-1.5cm}

    \captionof{figure}{
\textbf{Overview of the proposed regularization framework.}
$\mathcal{L}_{\mathrm{IntDrift}}$ and $\mathcal{L}_{\mathrm{Feat}}$ preserve functional behavior within protected hypercubes corresponding to previously learned tasks (pink). $\mathcal{L}_{\mathrm{Var}}$ encourages compact activation regions for newly acquired tasks (orange), while $\mathcal{L}_{\mathrm{Align}}$ limits excessive expansion of the cumulative feature space by aligning task centers across tasks. Together, these constraints produce compact and stable feature regions, illustrated by the purple contours.
}
    \label{fig:loss}
\end{figure*}

The effectiveness of the proposed constraints additionally depends on maintaining compact activation regions. If task representations become excessively dispersed, the cumulative hypercube expands unnecessarily, reducing the free space available for future adaptation. To encourage compactness, we minimize the intra-task variance around the batch centroid $\boldsymbol{\mu}^{(t)}
=
\frac{1}{N_B}
\sum_{i=1}^{N_B}
\mathbf{z}_i$,
using
\begin{equation}
\label{eq:reg_var}
\mathcal{L}_{\mathrm{Var}}
=
\frac{\lambda_{\mathrm{Var}}}{N_B \, d}
\sum_{i=1}^{N_B}
\left\|
\mathbf{z}_i
-
\boldsymbol{\mu}^{(t)}
\right\|_2^2,
\end{equation}
where $\lambda_{\mathrm{Var}}$ controls the strength of the compactness regularization. In Appendix~\ref{para:hypercube_growth}, we additionally analyze the resulting hypercube volumes and how much of the cumulative hypercube is occupied by the task-specific hypercubes.

Finally, even when individual task representations are compact, a single cumulative hypercube may grow excessively if consecutive tasks occupy distant regions of the feature space. To mitigate this expansion, we introduce an inter-task alignment objective that encourages the center $\mathbf{c}_t$ of the current batch to move toward the center $\mathbf{c}_{t-1}$ of the historical hypercube:
\begin{equation}
\label{eq:inter_task_alignment_loss}
\mathcal{L}_{\mathrm{Align}}
=
\lambda_{\mathrm{Align}}
\frac{
\left\|
\mathbf{m}
\odot
\big(
\mathbf{c}_{t}
-
\mathbf{c}_{t-1}
\big)
\right\|_2^2
}{
r_{t-1} + \eta
},
\end{equation}
where $\mathbf{c}_{t} = \frac{1}{2}(\min_i \mathbf{z}_i + \max_i \mathbf{z}_i)$ denotes the element-wise mid-range of the current batch, and $m_j=1$ if the interval spanned by the current batch along dimension $j$ does not overlap the corresponding historical interval $[\underline{z}_j, \overline{z}_j]$, and $0$ otherwise. The symbol $\odot$ denotes element-wise multiplication. For the cumulative hypercube $\mathcal{H}^{(\leq t-1)} = [\underline{\mathbf{z}}, \overline{\mathbf{z}}]$, its center and average radius are defined as $\mathbf{c}_{t-1} = \frac{1}{2}(\underline{\mathbf{z}} + \overline{\mathbf{z}})$ and $r_{t-1} = \frac{1}{2d}\sum_{j=1}^{d}(\overline{z}_j - \underline{z}_j)$, respectively. The constant $\eta > 0$ ensures numerical stability, while $\lambda_{\mathrm{Align}}$ controls the strength of the regularization. Together, $\mathcal{L}_{\mathrm{Var}}$ and $\mathcal{L}_{\mathrm{Align}}$ encourage compact task-specific representations and alignment with the historical feature region, limiting unnecessary hypercube expansion while preserving the flexibility to adapt to new tasks.

\paragraph{Composite Training Objective.}

The final optimization objective combines the task-specific cross-entropy (CE) loss with the proposed stability and geometric regularization terms:
\begin{equation*}
    \mathcal{L}_{\mathrm{Total}}
    =
    \mathcal{L}_{\mathrm{CE}}
    +
    \underbrace{
        \mathcal{L}_{\mathrm{IntDrift}}
        +
        \mathcal{L}_{\mathrm{Feat}}
    }_{\text{Functional Stability}}
    +
    \underbrace{
        \mathcal{L}_{\mathrm{Var}}
        +
        \mathcal{L}_{\mathrm{Align}}
    }_{\text{Geometric Structuring}}.
\end{equation*}

The interaction between these components is illustrated in Fig.~\ref{fig:loss}. The complete training procedure is provided in Appendix~\ref{appendix:sec_training_algorithm}.

\begin{table*}[t]
\centering
\caption{\textbf{Benchmarking against the SOTA.} We report AA across four diverse DIL benchmarks as the mean over ten runs for \our{} and over three runs for the baselines. \our{} achieves the highest average performance, with the largest gains on ImageNet-R and ImageNet-Mix. On DomainNet and ImageNet-C, \our{} remains highly competitive, matching or slightly exceeding the best baseline results.
 Presented results are achieved by integrating \our{} into KA-Prompt. Baseline results are sourced from~\cite{xu2025comp}.}
\label{tab:main_results}
\setlength{\tabcolsep}{3.8pt}
\begin{tabular}{llcccc>{\columncolor{avgcolor}}c}
\toprule
Method & Ref. & DomainNet & ImageNet-R & ImageNet-C & ImageNet-Mix & \textbf{Average} \\
\midrule
EWC & \textit{\scriptsize PNAS 2017} & $\res{47.10}{0.58}$ & $\res{51.83}{0.33}$ & $\res{65.20}{1.23}$ & $\res{57.82}{0.90}$ & $\res{55.49}{0.42}$ \\
LwF & \textit{\scriptsize T-PAMI 2017} & $\res{54.85}{0.06}$ & $\res{57.11}{0.95}$ & $\res{69.01}{1.21}$ & $\res{65.27}{0.93}$ & $\res{61.56}{0.45}$ \\
L2P & \textit{\scriptsize CVPR 2022} & $\res{53.74}{0.04}$ & $\res{56.55}{0.33}$ & $\res{77.86}{0.44}$ & $\res{64.20}{0.34}$ & $\res{63.09}{0.16}$ \\
S-Prompts & \textit{\scriptsize NeurIPS 2022} & $\res{44.08}{0.05}$ & $\res{27.23}{0.16}$ & $\res{60.25}{0.28}$ & $\res{24.25}{0.19}$ & $\res{38.95}{0.09}$ \\
DualPrompt & \textit{\scriptsize ECCV 2022} & $\res{55.18}{0.02}$ & $\res{59.47}{1.00}$ & $\res{78.52}{0.30}$ & $\res{63.57}{0.17}$ & $\res{64.19}{0.26}$ \\
ESN & \textit{\scriptsize AAAI 2023} & $\res{45.74}{0.25}$ & $\res{16.39}{0.23}$ & $\res{68.34}{0.39}$ & $\res{14.99}{0.53}$ & $\res{36.37}{0.19}$ \\
CODA-P & \textit{\scriptsize CVPR 2023} & $\res{55.81}{0.03}$ & $\res{55.21}{0.33}$ & $\res{78.25}{0.16}$ & $\res{64.92}{0.04}$ & $\res{63.55}{0.09}$ \\
InfLoRA & \textit{\scriptsize CVPR 2024} & $\res{48.76}{0.30}$ & $\res{41.20}{1.65}$ & $\res{53.12}{0.16}$ & $\res{35.38}{0.27}$ & $\res{44.62}{0.43}$ \\
Cprompt & \textit{\scriptsize CVPR 2024} & $\res{52.88}{0.12}$ & $\res{59.48}{1.09}$ & $\res{70.08}{0.07}$ & $\res{63.64}{0.74}$ & $\res{61.52}{0.33}$ \\
C-Prompt & \textit{\scriptsize IJCV 2024} & $\res{58.66}{0.05}$ & $\res{62.43}{0.49}$ & $\res{79.84}{0.38}$ & $\res{65.35}{0.52}$ & $\res{66.57}{0.20}$ \\
KA-Prompt & \textit{\scriptsize ICML 2025} & $\underline{\res{62.91}{0.14}}$ & $\underline{\res{66.51}{0.36}}$ & $\underline{\res{85.43}{0.56}}$ & $\underline{\res{70.35}{0.32}}$ & $\underline{\res{71.30}{0.19}}$ \\
\midrule
\textbf{\our{}} & \textit{\scriptsize This Paper} & $\mathbf{\res{62.96}{0.09}}$ & $\mathbf{\res{67.90}{0.44}}$ & $\mathbf{\res{85.75}{0.31}}$ & $\mathbf{\res{70.94}{0.41}}$ & $\mathbf{\res{71.89}{0.17}}$ \\
\bottomrule
\end{tabular}
\end{table*}

\section{Experiments}
\label{sec:experiments}

We evaluate \our{} in the DIL setting, focusing on three key areas: comparative performance against SOTA methods, plasticity under distribution shifts, and the individual contribution of each loss component to the stability-plasticity trade-off.

\subsection{Experimental Setup}

\paragraph{Datasets and Baselines.}
We evaluate \our{} on four standard continual learning benchmarks: DomainNet, ImageNet-R, ImageNet-C, and ImageNet-Mix. The proposed method is compared against classical regularization-based baselines (EWC, LwF) as well as recent SOTA approaches, including L2P, DualPrompt, CODA-Prompt, KA-Prompt, and InfLoRA, with particular emphasis on prompt-based continual learning methods. We report the Average Accuracy (AA) metric; its formal definition is provided in Appendix~\ref{appendix:sec_metrics}.

\paragraph{Implementation Details.}
For fair comparison, we follow KA-Prompt in using the same task splits, continual learning protocols, and ViT-B/16 backbone pre-trained on ImageNet-21k without ImageNet-1k fine-tuning~\cite{dosovitskiy2021imageworth16x16words}. Results for \our{} are averaged over ten runs with different random seeds. For experiments combining CODA-Prompt with \our{}, including ablations and sensitivity analyses, we use a ViT-B/16 backbone additionally fine-tuned on ImageNet-1k~\cite{Steiner2021HowTT}. Further implementation details and hyperparameter configurations are provided in Appendix~\ref{appendix:sec_implementation_details}.

\subsection{Results}

\paragraph{Main Results.}
Table~\ref{tab:main_results} reports the AA across four DIL benchmarks. Overall, \our{} achieves the highest average performance among the compared methods. In particular, it establishes new state-of-the-art results on ImageNet-R and ImageNet-Mix, outperforming the strongest baseline, KA-Prompt, by \textbf{1.39 p.p.} and \textbf{0.59 p.p.}, respectively. \our{} also achieves the best result on ImageNet-C.

\begin{wrapfigure}{R}{0.48\textwidth}
    \vspace{-15pt}
    \includegraphics[width=\linewidth]{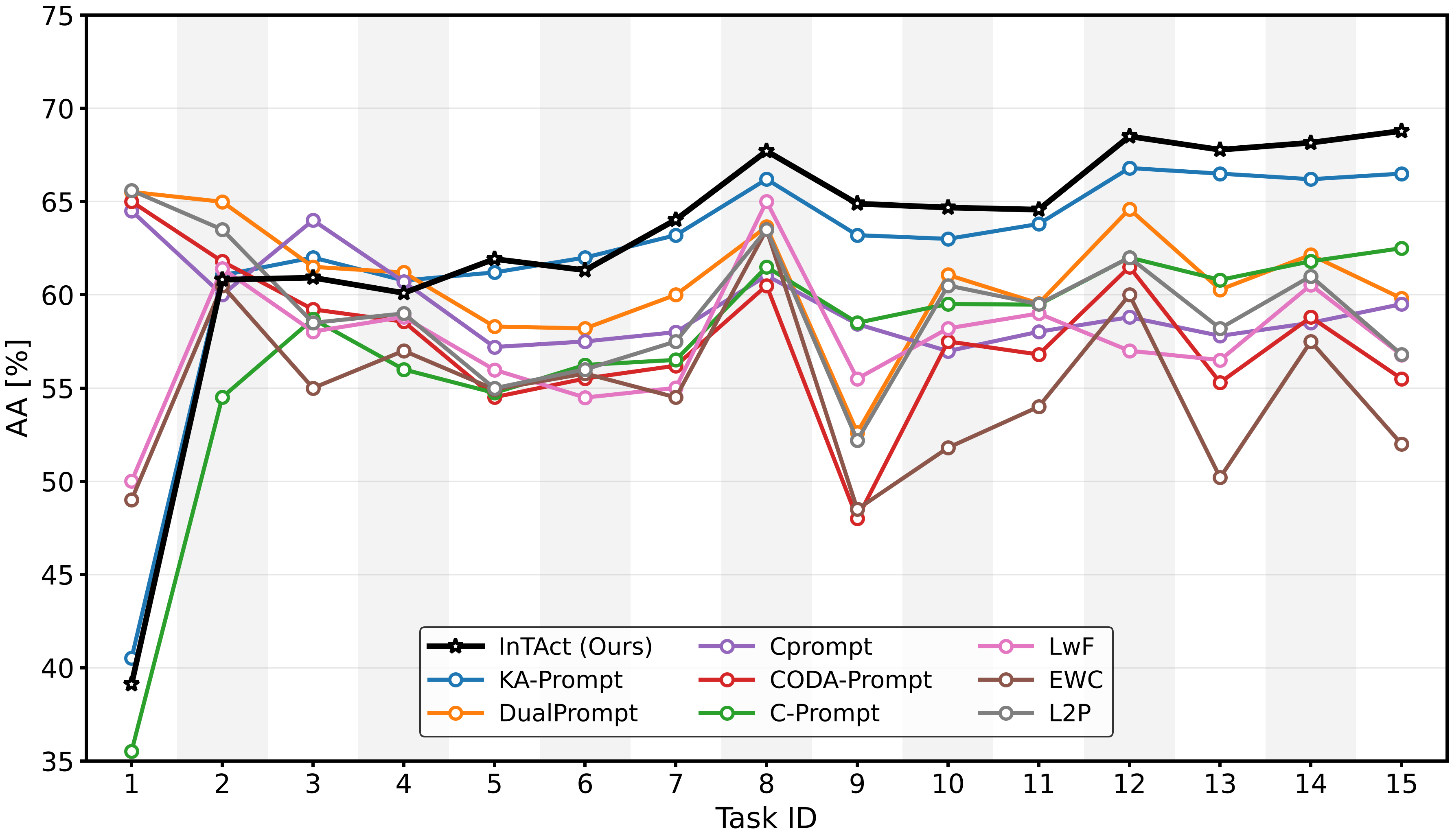}
    \caption{\textbf{AA Evolution on ImageNet-R.} Above plot reports the AA across 15 incremental tasks. \our{} (black line with stars) outperforms SOTA prompt-based baselines (e.g., KA-Prompt, DualPrompt) and regularization methods (e.g., EWC, LwF) from $t=7$ onward. Notably, our method maintains high accuracy even during challenging task transitions (e.g., $t=9$), where most other methods exhibit significant forgetting. Baseline results are sourced from~\cite{xu2025comp}.}
    \label{appendix:fig_imr_comp}
    \vspace{-30pt}
\end{wrapfigure}

On DomainNet, \our{} performs on par with KA-Prompt. Importantly, \our{} is integrated into KA-Prompt without modifying its prompt selection or training procedures, demonstrating that the proposed regularization can improve or match the original architecture across all evaluated benchmarks. This indicates that the observed gains arise from the proposed feature-space regularization rather than architecture-specific modifications.

These results suggest that parameter-space regularization may be insufficient to fully preserve feature representations under substantial domain shifts, as constraining weight updates does not necessarily prevent changes in the feature space. By directly regularizing functional changes within protected regions, \our{} offers an alternative way to promote stability while retaining flexibility for future adaptation. Although the gains on corruption-heavy benchmarks, particularly ImageNet-C and ImageNet-Mix, are more modest, they are consistent with the potential benefits of stabilizing feature representations under distribution shifts.

\paragraph{Integration with Other Prompt-Based Methods.} The gains over KA-Prompt are relatively modest, which may be partly explained by its exponential moving average of classifier weights, which already helps mitigate classifier drift. In contrast, methods without this mechanism benefit more substantially: integrating \our{} into L2P, DualPrompt, and CODA-Prompt yields improvements of $4.1$--$13.0$ p.p. (see Appendix~\ref{appendix:sec_intact_in_other_methods}).

\paragraph{AA Evolution Analysis on ImageNet-R.}
Fig.~\ref{appendix:fig_imr_comp} shows the evolution of AA on ImageNet-R over 15 sequential tasks. Since \our{} is built on KA-Prompt, it starts at a comparable performance level and surpasses all competing methods from $t=7$ onward. While classical baselines such as EWC and LwF exhibit a sharp performance drop around the 9th task, \our{} maintains a substantially more stable trajectory, with only minor degradation as new tasks are introduced. Furthermore, compared with recent prompt-based approaches, including KA-Prompt and C-Prompt, \our{} exhibits a more favorable balance between stability and plasticity. By the final task, the strongest baselines remain below $67\%$ AA.

\subsection{Ablation Study}
\label{sec:ablation}
To assess the contribution of each component in \our{}, we conduct a systematic ablation study on the ImageNet-R benchmark (Table~\ref{tab:ablation_results}) using CODA-Prompt + \our{}; the absolute AA therefore differs from Tab.~\ref{tab:main_results}, which uses KA-Prompt and a different backbone. Specifically, we remove one regularization term at a time to isolate its role in the stability--plasticity trade-off.

The results demonstrate that all components contribute to the final performance, with the interval drift constraint playing the most critical role. Removing the proposed interval-based regularization term ($\mathcal{L}_{\mathrm{IntDrift}}$) causes the largest degradation, reducing AA by \textbf{3.48} p.p. This confirms that directly constraining functional drift within protected feature regions is the primary mechanism responsible for preserving previously learned knowledge across domain shifts.

The feature consistency term ($\mathcal{L}_{\mathrm{Feat}}$) is also important, with its removal decreasing performance by \textbf{1.79} p.p. This indicates that stabilizing intermediate representations remains necessary even when the classifier behavior is explicitly constrained.

The remaining geometric regularization terms further improve the structure of the feature space. Removing the compactness objective ($\mathcal{L}_{\mathrm{Var}}$) lowers AA by \textbf{0.95} p.p., suggesting that unconstrained feature dispersion leads to larger protected regions and increased inter-task interference. Similarly, excluding the alignment term ($\mathcal{L}_{\mathrm{Align}}$) results in a \textbf{0.47} p.p. drop, indicating that aligning task centers helps limit unnecessary expansion of the cumulative hypercube and preserves future plasticity. An extension of Tab.~\ref{tab:ablation_results} to all pairwise removals of the four terms is provided in Tab.~\ref{tab:pairwise_ablation} in Appendix~\ref{appendix:additional_results}; the strongest interaction occurs between $\mathcal{L}_{\mathrm{IntDrift}}$ and $\mathcal{L}_{\mathrm{Var}}$, whose joint removal causes a drop nearly twice the sum of their individual drops. In contrast, the gradients of $\mathcal{L}_{\mathrm{IntDrift}}$ and $\mathcal{L}_{\mathrm{Align}}$ remain nearly orthogonal during training (average cosine similarity of at most $0.012$; Appendix~\ref{appendix:additional_results}), indicating that the alignment term does not work against the interval drift constraint.

\begin{wraptable}{R}{0.48\textwidth}
    \centering
    \caption{\textbf{Ablation Study.} Impact of removing individual loss components of CODA-Prompt + \our{} on ImageNet-R. We report the performance drop ($\Delta$) relative to the baseline \our{}. The results are averaged over 3 random seeds.}
    \label{tab:ablation_results}
    
    \adjustbox{max width=\linewidth}{
    \begin{tabular}{ccccc>{\columncolor{avgcolor}}c}
    \toprule
    $\mathcal{L}_{\mathrm{IntDrift}}$ & $\mathcal{L}_{\mathrm{Var}}$ & $\mathcal{L}_{\mathrm{Align}}$ & $\mathcal{L}_{\mathrm{Feat}}$ & \textbf{AA (\%)} & \textbf{$\Delta$} \\
    \midrule
    \checkmark & \checkmark & \checkmark & \checkmark & \textbf{74.23} & --- \\
    \midrule
    $\times$ & \checkmark & \checkmark & \checkmark & 70.75 & -3.48 \\
    \checkmark & $\times$ & \checkmark & \checkmark & 73.28 & -0.95 \\
    \checkmark & \checkmark & $\times$ & \checkmark & 73.76 & -0.47 \\
    \checkmark & \checkmark & \checkmark & $\times$ & 72.44 & -1.79 \\
    \bottomrule
    \end{tabular}
    }
    \vspace{-15pt} 
\end{wraptable}

\begin{figure*}[t]
  \centering
  \begin{subfigure}[b]{0.32\linewidth}
    \centering
    \includegraphics[width=\linewidth]{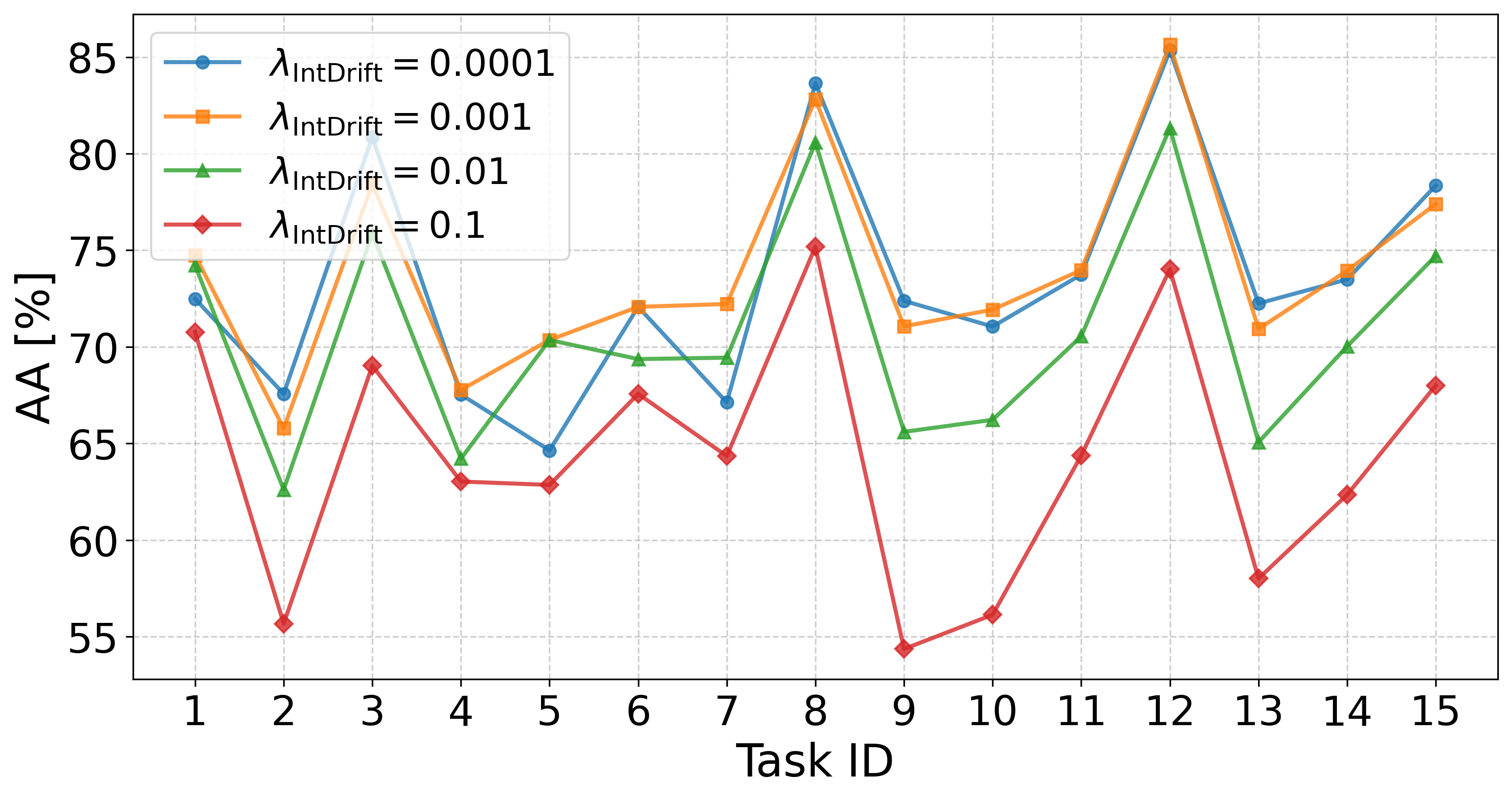}
    \caption{Sensitivity to $\lambda_{\mathrm{IntDrift}}$ changes}
    \label{fig:sens_intdrift}
  \end{subfigure}
  \hfill
  \begin{subfigure}[b]{0.32\linewidth}
    \centering
    \includegraphics[width=\linewidth]{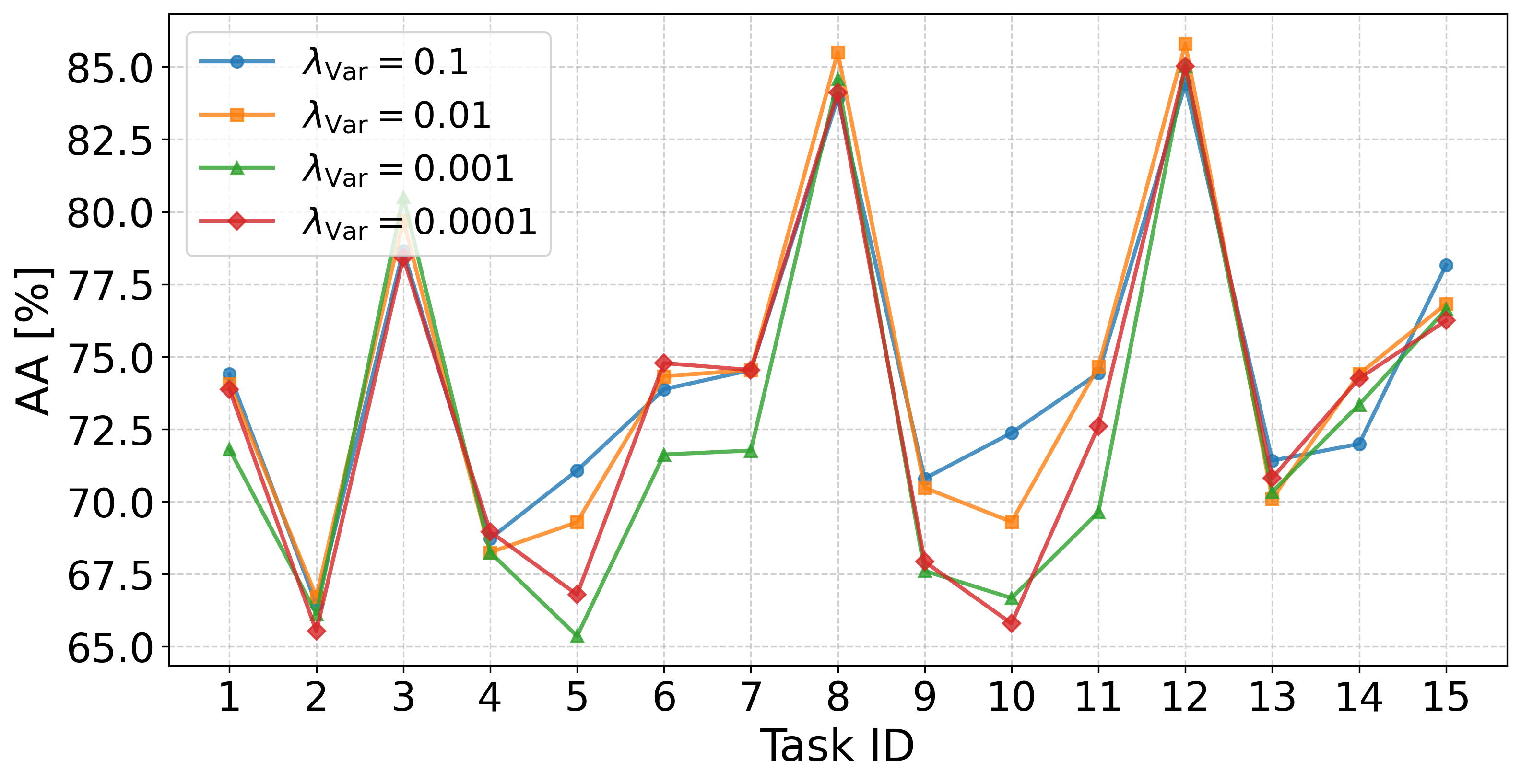}
    \caption{Sensitivity to $\lambda_{\mathrm{Var}}$ changes}
    \label{fig:sens_var}
  \end{subfigure}
  \hfill
  \begin{subfigure}[b]{0.32\linewidth}
    \centering
    \includegraphics[width=\linewidth]{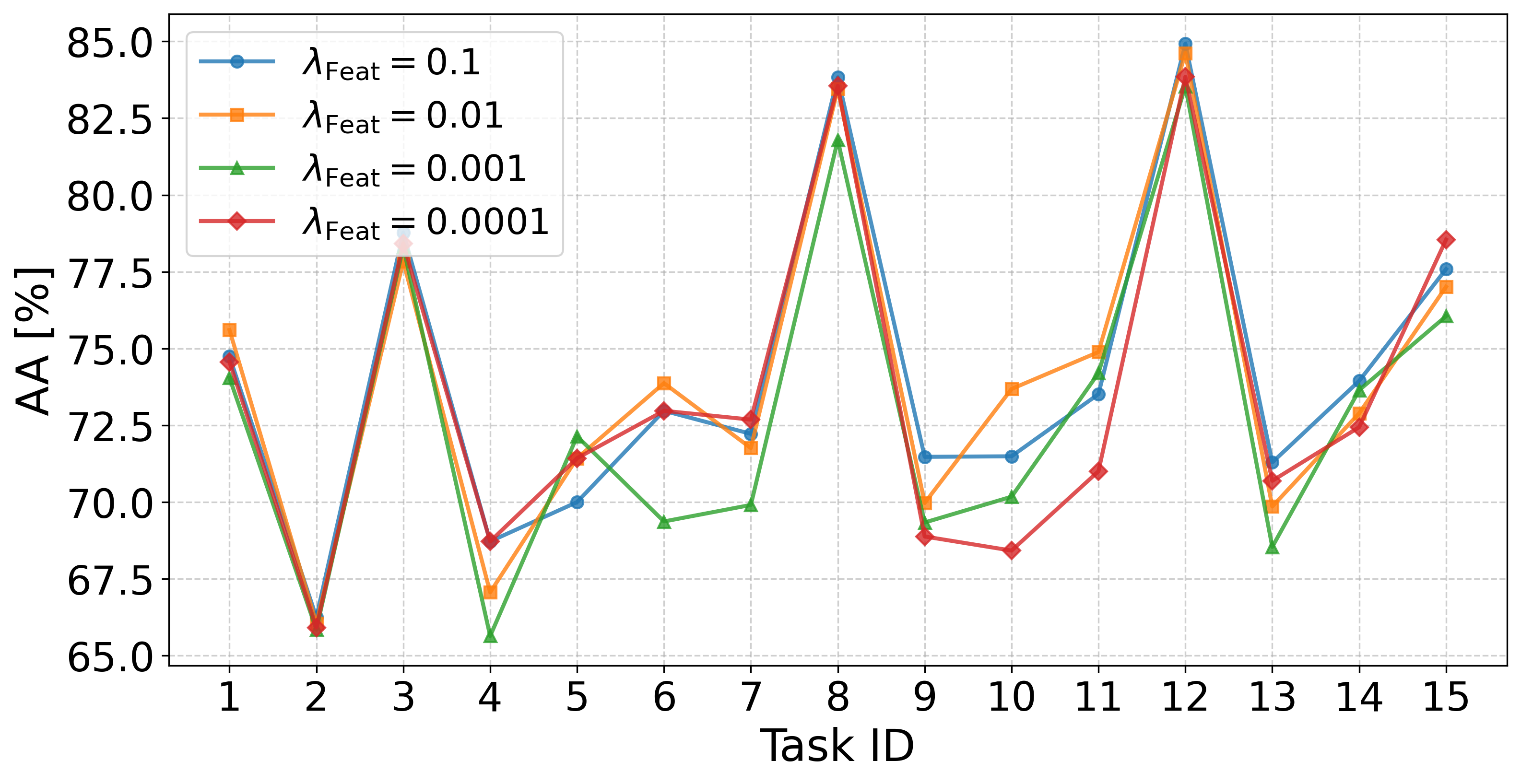}
    \caption{Sensitivity to $\lambda_{\mathrm{Feat}}$ changes}
    \label{fig:sens_feat}
  \end{subfigure}
  \vspace{-2mm}
  \caption{Hyperparameter sensitivity of CODA-Prompt + \our{} on ImageNet-R: final accuracy on each task, averaged over three independent runs. 
(a) Lower drift penalties preserve classifier plasticity, allowing adaptation to new tasks. 
(b) Stronger variance regularization promotes more compact representations and improves stability. 
(c) A sufficiently strong feature consistency penalty is crucial for preventing drift in the backbone's representations.}
  \label{fig:sensitivity_analysis}
  \vspace{-3mm}
\end{figure*}

\subsection{Sensitivity Analysis on ImageNet-R}
\label{sec:sensitivity}

We analyze the sensitivity of \our{} to its main regularization hyperparameters on the ImageNet-R benchmark, using CODA-Prompt + \our{}.

\paragraph{Internal Drift Regularization ($\lambda_{\mathrm{IntDrift}}$).}
Fig.~\ref{fig:sens_intdrift} shows that smaller values of $\lambda_{\mathrm{IntDrift}}$ yield higher accuracy, while stronger regularization degrades performance. This highlights the need to balance stability and plasticity: although functional drift must be controlled, the classifier needs sufficient flexibility to adapt its decision boundaries to the substantial domain shifts in ImageNet-R.

\paragraph{Variance Regularization ($\lambda_{\mathrm{Var}}$).}
As illustrated in Fig.~\ref{fig:sens_var}, increasing $\lambda_{\mathrm{Var}}$ generally improves performance by encouraging more compact feature representations. By reducing the overall volume of the cumulative hypercube, this regularization limits unnecessary expansion of the protected region, thereby reducing inter-task interference and preserving feature-space capacity for future adaptation.

\paragraph{Feature Stability ($\lambda_{\mathrm{Feat}}$).}
Fig.~\ref{fig:sens_feat} shows that stronger feature consistency regularization generally improves accuracy, although not monotonically. By penalizing deviations from the frozen model's representations, $\mathcal{L}_{\mathrm{Feat}}$ mitigates representational drift caused by prompt updates, helping preserve the mapping of historical inputs to their protected feature regions.

\subsection{Extended Analysis}
Appendix~\ref{appendix:additional_results} extends the evaluation of \our{} with CIL experiments, sensitivity to the hypercube coverage parameter ($\alpha$), and analyses of historical activation coverage, hypercube compactness, and occupancy over long task sequences.

\section{Conclusion}
\label{sec:conclusions}

We introduced \our{}, a continual learning framework that mitigates catastrophic forgetting by directly constraining functional behavior in feature space rather than regularizing individual model parameters. Leveraging IA, the proposed method preserves stability within regions associated with previously learned tasks while maintaining flexibility in unconstrained regions of the representation space. This explicit separation between stability and plasticity enables effective continual adaptation without relying on replay buffers or architectural expansion. Extensive experiments demonstrate that \our{} consistently improves existing prompt-based continual learning methods and achieves SOTA performance in the DIL setting.

\paragraph{Limitations.}

Our theoretical guarantee provides a local, layer-wise upper bound on functional drift, conditional on the internal drift loss being sufficiently small and historical representations remaining within the protected hypercube. Consequently, the strength of the guarantee depends on the effectiveness of the optimization and the geometry of the protected region. Larger hypercubes cover a broader portion of the feature space but may yield looser bounds, whereas smaller hypercubes provide tighter guarantees over more limited regions. Furthermore, our approach relies on hypercubes to capture regions occupied by historical representations, which may not fully reflect the geometry of complex feature distributions. Exploring alternative representations of historical regions that provide a more flexible characterization of the feature space is an important direction for future work.

\paragraph{Acknowledgments.}
The work of P. Spurek and P. Krukowski was supported by the National Centre of Science (Poland) Grant No. 2023/50/E/ST6/00068. We gratefully acknowledge Polish high-performance computing infrastructure PLGrid (HPC Centers: ACK Cyfronet AGH, WCSS, CI TASK) for providing computer facilities and support within computational grant no. PLG/2026/019272.

\bibliographystyle{unsrt}
\bibliography{ref}

\appendix

\section{Extended Overview of Related Works}
\label{appendix:sec_related_work}

This section reviews classical and recent approaches to continual learning. We first introduce the three main paradigms: replay-based, regularization-based, and parameter-isolation methods. We then discuss parameter-efficient approaches, focusing on prompt-based continual learning, followed by methods addressing representational drift in feature space. Finally, we review applications of interval arithmetic (IA) to continual learning.

\paragraph{Continual Learning.}
Continual learning methods~\cite{2019parisi+4,2021delange+7,2022masana+5} can generally be divided into three categories. \emph{Replay-based} methods use memory or rehearsal mechanisms to retain information from previous tasks. These include exemplar replay, which stores selected training samples~\cite{2018chaudhry+3,2018riemer+6,2019chaudhry+6,2020buzzega+4,lopezpaz2022gradientepisodicmemorycontinual,2024liang+1,2025urettini+1}, and generative replay, which synthesizes previously observed data using generative models~\cite{2017shin+3,2018wu+4}. \emph{Regularization-based} methods constrain parameter updates to preserve previously acquired knowledge. Such constraints can be derived from current-task data~\cite{li2017learningforgetting,2023kim+3} or information accumulated from previous tasks~\cite{Kirkpatrick_2017,zenke2017continuallearningsynapticintelligence,aljundi2018memoryawaresynapseslearning,2025chen+12}. Other approaches restrict weight updates to the null space of previous-task feature covariance~\cite{2022kong+3,2021tang+4,2021wang+3,2022wang+4}, while~\cite{2024liang+1} uses gradient information from previous tasks to construct a subspace for the LoRA dimensionality-reduction matrix. Finally, \emph{parameter-isolation} methods allocate task-specific parameters or subnetworks to prevent interference. Examples include Progressive Neural Networks~\cite{2016rusu+7,rusu2022progressiveneuralnetworks}, Piggyback~\cite{2018mallya+2}, PackNet~\cite{2018mallya+1}, HAT~\cite{2018serra+3}, SupSup~\cite{2020wortsman+6}, ESN~\cite{wang2023ESN}, and InfLoRA~\cite{liang2024InfLoRA}. Although effective in task-aware settings, many such methods require task identification at inference time.

\paragraph{Prompt-Based Continual Learning.}
Prompt-based continual learning adapts a frozen pre-trained backbone using a small set of learnable parameters called \emph{prompts}. Typically, these are vectors prepended to input embeddings or inserted into intermediate layers, steering the model toward new tasks without modifying the backbone weights. A key development is the use of a \emph{prompt pool}. L2P~\cite{wang2022L2P} maintains a shared set of prompt vectors and selects a sparse subset by matching input features to learned prompt keys. This enables instance-wise prompt selection at inference time without requiring task identity. Subsequent methods introduce additional structure and constraints on prompt learning. DualPrompt~\cite{wang2022DualPrompt} separates prompts into general components shared across tasks and task-specific components that capture task-dependent knowledge. CODA-Prompt~\cite{smith2023CODAPrompt} employs decomposed attention to dynamically combine prompt components, aiming to reduce interference between tasks. Compositional Prompting (C-Prompt)~\cite{Liu2024C_Prompt} promotes the sharing of compositional prompt components across domains, while Consistent Prompting (CPrompt)~\cite{gao2024CPrompt} encourages consistency between task-specific prompts to mitigate representational drift. More recently, KA-Prompt~\cite{xu2025comp} addresses component-wise misalignment between prompts learned for different domains by initializing new prompts from relevant historical ones and encouraging consistency with them during training. Despite their effectiveness in parameter-efficient, rehearsal-free adaptation, prompt-based methods can still suffer from representational drift and classifier instability, particularly in domain-incremental learning (DIL). Although the backbone remains frozen, prompt updates alter the representations passed to the classifier, while substantial distribution shifts may require adapting its decision boundaries. Thus, preserving pre-trained weights alone does not ensure stable predictions across domains.

\paragraph{Feature Stability in Continual Learning.}
A related line of research addresses the drift of previously learned representations as new tasks are introduced. Many approaches rely on class prototypes~\cite{2022janoson+3,goswami2024fecamexploitingheterogeneityclass} and aim to preserve their positions in feature space. SDC~\cite{2020yu+7} and~\cite{2022toldo+1} estimate and compensate for feature drift using current-task data after each training phase. NAPA-VQ~\cite{2023malepathirana+2} and Prototype Reminiscence~\cite{2023shi+1} leverage topological information to reshape historical prototypes, while FeTrIL~\cite{2023petit+4} translates features to align old and new class distributions. ADC~\cite{2024goswami+5} generates adversarial pseudo-exemplars of new classes to adjust previous prototypes, and EASE~\cite{2024zhou+3} uses semantic-guided prototype complementation to update old-class representations. Another approach~\cite{2025liu+1} constructs a drift-resistant feature space in which model weights can adapt to new tasks while limiting interference with previously learned representations. More broadly, representation-level regularization seeks to constrain changes in intermediate activations rather than directly restricting parameter updates. Such approaches are closely related to activation-based methods that preserve task-specific regions of feature space, although efficiently characterizing and maintaining these regions over long task sequences remains challenging.

\paragraph{Interval-Based Methods in Continual Learning.}
IA~\cite{dahlquist2008numerical,moore2009introduction} has also been explored in continual learning. InterContiNet~\cite{wolczyk2022continual} constrains network weights to task-specific intervals, whose intersection defines configurations that maintain performance across tasks. HINT~\cite{2025krukowski+5} instead constructs intervals in the embedding space and uses a hypernetwork to map them to network weights. Together, these methods explore how interval representations can preserve regions of parameter or feature space associated with previously learned tasks.

\section{Preliminaries}
\label{appendix:sec_preliminaries}

\paragraph{Prompt-Based Continual Learning.}
Prompt-based continual learning exploits the representational capacity of pretrained ViT~\cite{dosovitskiy2021imageworth16x16words} by introducing a small set of learnable parameters (prompts) to adapt the model to new tasks while keeping the backbone frozen. This strategy avoids the computational cost and catastrophic forgetting associated with fine-tuning the entire network.

Formally, consider a ViT encoder $f_{\boldsymbol{\theta}}$ with fixed parameters $\boldsymbol{\theta}$. The input to the $l$-th layer is denoted as $\mathbf{z}^{(l)} \in \mathbb{R}^{L \times D}$, where $L$ is the sequence length and $D$ is the embedding dimension. To adapt the representation, standard approaches employ Prefix-Tuning~\cite{li2021prefixtuningoptimizingcontinuousprompts}. This mechanism prepends learnable vectors to the internal attention layers. Specifically, learnable prompt matrices $\mathbf{P}_K, \mathbf{P}_V \in \mathbb{R}^{L_p \times D}$ (where $L_p$ is the prompt length) are defined and concatenated with the standard Key and Value projections:
\begin{equation}
\label{eq:prefix}
\begin{aligned}
    \mathbf{Q} &= \mathbf{z}^{(l)} \mathbf{W}_Q, \\
    \mathbf{K}' &= \left[ \mathbf{P}_K ; \mathbf{z}^{(l)} \mathbf{W}_K \right], \\
    \mathbf{V}' &= \left[ \mathbf{P}_V ; \mathbf{z}^{(l)} \mathbf{W}_V \right], \\
    \mathbf{z}^{(l+1)} &= \mathrm{MSA}(\mathbf{Q}, \mathbf{K}', \mathbf{V}'),
\end{aligned}
\end{equation}
where $\mathrm{MSA}$ denotes Multi-Head Self-Attention, $[;]$ represents concatenation along the sequence dimension, and $\mathbf{W}_Q, \mathbf{W}_K, \mathbf{W}_V$ are the frozen projection matrices. In this framework, only the prompts $\mathbf{P}_K, \mathbf{P}_V$ and the final classification head are optimized, allowing the system to adapt to non-stationary data streams efficiently.

\paragraph{Prompt Selection and Memory Structure.}
A central challenge in task-agnostic continual learning is selecting appropriate prompts without access to task identifiers at test time. To address this, methods like Learning to Prompt (L2P)~\cite{wang2022L2P} maintain a shared prompt pool $\mathcal{P} = \{(\mathbf{k}_i, \mathbf{p}_i)\}_{i=1}^M$, where each learnable prompt $\mathbf{p}_i$ is paired with a key $\mathbf{k}_i \in \mathbb{R}^{D}$.

The inference process operates via a content-based retrieval mechanism. First, the input $\mathbf{x}$ is passed through the frozen encoder to generate a query vector $\mathbf{q}(\mathbf{x})$ (typically the \texttt{[CLS]} embedding). This query is matched against the stored keys $\{\mathbf{k}_i\}$ using cosine similarity to identify the subset of the $K$ most relevant prompts. These selected prompts are then explicitly concatenated with the input embeddings or attention prefixes, effectively conditioning the frozen backbone on the retrieved context. During training, optimization is sparse; gradients are applied only to the selected prompt-key pairs, allowing the remaining pool contents to preserve knowledge from disparate tasks.

\paragraph{Optimization Objective.}
Since the discrete retrieval of keys (Top-$K$) is non-differentiable, gradients cannot backpropagate to the keys directly. Consequently, prompt-based methods typically minimize a composite objective $\mathcal{L}_{\mathrm{Total}} = \mathcal{L}_{\mathrm{task}} + \lambda \mathcal{L}_{\mathrm{match}}$. Here, $\mathcal{L}_{\mathrm{task}}$ (e.g., Cross-Entropy) updates the prompts and classifier head, while the surrogate loss $\mathcal{L}_{\mathrm{match}}$ pulls the keys $\{\mathbf{k}_i\}$ closer to the queries of their assigned inputs, ensuring that the retrieval mechanism aligns with the data distribution.

\paragraph{Hierarchical and Decomposed Prompting.}
DualPrompt~\cite{wang2022DualPrompt} decouples plasticity and stability by partitioning the prompts into distinct functional groups: G-Prompts ($\mathbf{g}$) for task-invariant features and E-Prompts ($\mathbf{e}_t$) for task-specific experts. These are inserted into complementary depths of the backbone:
\begin{equation}
\label{eq:dualprompt}
\mathbf{z}^{(l+1)} = \begin{cases}
      \mathrm{MSA}(\mathbf{z}^{(l)}, \mathbf{g}) & \text{if } l \leq L_g \quad \text{(General Layers)} \\
      \mathrm{MSA}(\mathbf{z}^{(l)}, \mathbf{e}_t) & \text{if } L_g < l \leq L_e \quad \text{(Expert Layers)}
   \end{cases}
\end{equation}
where $\mathbf{e}_t$ is retrieved via a task-specific key $\mathbf{k}_t$, and the remaining layers are not prompted.

CODA-Prompt~\cite{smith2023CODAPrompt} generalizes this by replacing discrete selection with a soft, weighted summation of components. The input query $\mathbf{q}(\mathbf{x})$ attends to the entire pool of prompt components $\{\mathbf{P}_m\}_{m=1}^M$, each paired with a key $\mathbf{k}_m$ and an attention vector $\mathbf{A}_m$, via a learnable attention mechanism:
\begin{equation}
\label{eq:coda_attn}
\alpha_m(\mathbf{x}) = \cos\left( \mathbf{q}(\mathbf{x}) \odot \mathbf{A}_m, \mathbf{k}_m \right), \quad \mathbf{p}(\mathbf{x}) = \sum_{m=1}^M \alpha_m(\mathbf{x}) \mathbf{P}_m.
\end{equation}
Here, $\cos(\cdot, \cdot)$ denotes the cosine similarity, $\mathbf{A}_m$ acts as a query-modulation vector, and $\odot$ denotes the Hadamard product. This differentiable formulation allows the prompt $\mathbf{p}(\mathbf{x})$ to form complex linear combinations of existing knowledge, stabilized by an orthogonality penalty on the components, keys, and attention vectors to minimize inter-component redundancy.
\section{Implementation Details}
\label{appendix:sec_implementation_details}

\paragraph{Architecture and Training.}
All prompt-based experiments use a ViT-B/16 backbone~\cite{dosovitskiy2021imageworth16x16words} pre-trained on ImageNet-21k, with the exact weights depending on the underlying codebase. Experiments based on KA-Prompt, including the main results and occupancy analysis, use the original ImageNet-21k weights without ImageNet-1k fine-tuning. Experiments based on the CODA-Prompt codebase, including ablations, sensitivity analyses, and comparisons with other prompt methods, use weights additionally fine-tuned on ImageNet-1k~\cite{Steiner2021HowTT}. L2P + \our{} and DualPrompt + \our{} follow the latter configuration. In all experiments, the backbone remains frozen, and adaptation is performed by optimizing the prompts and shared classification head. Since we consider DIL, all tasks share the same label space and classifier output dimension. The model must therefore adapt to domain shifts while preserving consistent decision boundaries across tasks.

\paragraph{Datasets and Protocols.}
We evaluate \our{} on four benchmarks designed to assess robustness against varying degrees of domain shift:
\textbf{DomainNet}~\cite{peng2019moment} offers the most severe visual shifts, consisting of six distinct domains (Clipart, Infograph, Painting, Quickdraw, Real, Sketch). In our DIL setting, each domain constitutes a separate task, processed sequentially.
\textbf{ImageNet-R}~\cite{hendrycks2021many} contains 200 classes rendered in diverse artistic styles (e.g., cartoons, embroidery). To construct a DIL sequence, we partition the dataset into 15 sequential tasks, ensuring all classes are present in every task but represented by different samples/styles.
\textbf{ImageNet-C}~\cite{hendrycks2019benchmarkingneuralnetworkrobustness} evaluates robustness against corruption. Following the protocol of \cite{Liu2024C_Prompt}, we utilize the 200 classes overlapping with ImageNet-R, subjecting them to 15 types of algorithmic perturbations (e.g., blur, noise) to form the task sequence.
Finally, \textbf{ImageNet-Mix}~\cite{Liu2024C_Prompt} combines ImageNet-R styles with ImageNet-C corruptions, resulting in a 30-task benchmark that rigorously tests stability under compound distributional shifts. All datasets are normalized and transformed during both training and evaluation, following the KA-Prompt method~\cite{xu2025comp}.

\paragraph{Code and Reproducibility.}
For the main experiments (Tab.~\ref{tab:main_results}), we integrate \our{} into the official KA-Prompt implementation~\cite{xu2025comp}, adopting its task splits and evaluation protocol. For experiments with L2P, DualPrompt, and CODA-Prompt, we use the official CODA-Prompt implementation and its corresponding task splits and evaluation protocol. In both cases, \our{} is integrated without modifying the original prompt selection or training procedures.

Experiments were conducted on NVIDIA H100 (80 GB), A100 (40 GB), and V100 (32 GB) GPUs within a DGX cluster, as well as standalone RTX 4090 (24 GB) machines. Training pipelines, configuration files, and reproduction scripts are available in our GitHub repository.

\section{Hyperparameter Configuration}
\label{appendix:selected_hyperparams}

For KA-Prompt + \our{}, we keep all hyperparameters of KA-Prompt, including its EMA classifier, fix $\lambda_{\mathrm{Align}} = 1$, and select the remaining coefficients via grid search over $\lambda_{\mathrm{Var}} \in \{10^{-2}, 10^{-1}, 1\}$, $\lambda_{\mathrm{IntDrift}} \in \{10^{-4}, 10^{-3}, 10^{-2}\}$, and $\lambda_{\mathrm{Feat}} \in \{10^{-2}, 10^{-1}\}$. For CODA-Prompt + \our{}, we use identical backbone and prompt configurations as the baseline, fix $\lambda_{\mathrm{Align}} = 1$, and search over $\lambda_{\mathrm{Var}} \in \{10^{-3}, \dots, 1.0\}$ and $\lambda_{\mathrm{IntDrift}}, \lambda_{\mathrm{Feat}} \in \{10^{-6}, \dots, 10^{-1}\}$. The specific values selected for the final evaluation are detailed in Table~\ref{appendix:tab_prompt_hp}.

\begin{table}[h]
\centering
\small
\caption{\textbf{Optimal Hyperparameters.} Selected hyperparameters for \our{} integrated into KA-Prompt and into CODA-Prompt across different benchmarks.}
\label{appendix:tab_prompt_hp}
\setlength{\tabcolsep}{10pt}
\begin{tabular}{lcccccc}
\toprule
 & \multicolumn{3}{c}{KA-Prompt + \our{}} & \multicolumn{3}{c}{CODA-Prompt + \our{}} \\
\cmidrule(lr){2-4} \cmidrule(lr){5-7}
Dataset & $\lambda_{\mathrm{Feat}}$ & $\lambda_{\mathrm{IntDrift}}$ & $\lambda_{\mathrm{Var}}$ & $\lambda_{\mathrm{Feat}}$ & $\lambda_{\mathrm{IntDrift}}$ & $\lambda_{\mathrm{Var}}$ \\
\midrule
DomainNet & $10^{-2}$ & $10^{-4}$ & $10^{-1}$ & $10^{-4}$ & $10^{-3}$ & $1.0$ \\
ImageNet-R & $10^{-1}$ & $10^{-3}$ & $1.0$ & $10^{-1}$ & $10^{-3}$ & $10^{-1}$ \\
ImageNet-C & $10^{-2}$ & $10^{-4}$ & $1.0$ & $10^{-1}$ & $10^{-4}$ & $10^{-1}$ \\
ImageNet-Mix & $10^{-2}$ & $10^{-4}$ & $1.0$ & $10^{-1}$ & $10^{-4}$ & $10^{-1}$ \\
\bottomrule
\end{tabular}
\end{table}

\section{Interval Arithmetic}\label{appendix:interval_arithmetic}

IA, introduced in its modern form by Ramon E. Moore~\cite{moore2009introduction}, provides a method for performing computations on ranges of real numbers rather than single point values. An interval $x$ is a closed, bounded set of real numbers defined by its endpoints, $x = [\underline{x}, \overline{x}]$, such that $\underline{x} \le \overline{x}$. This framework is essential for validated numerics, as it allows for the rigorous containment of numerical errors, including rounding errors and uncertainties in initial data.

\subsection{Fundamental Operations}

The fundamental principle of IA is the \emph{inclusion property}: the resulting interval of an operation must contain all possible results from applying the same operation to any real numbers within the operand intervals. Let $x = [\underline{x}, \overline{x}]$ and $y = [\underline{y}, \overline{y}]$ be two intervals. The elementary arithmetic operations are defined as follows:

\begin{description}
    \item[Addition.] The sum is obtained by adding the respective endpoints:
    \begin{equation*}
    x + y = [\underline{x} + \underline{y}, \overline{x} + \overline{y}]
    \end{equation*}
    
    \item[Subtraction.] The difference is computed by cross-adding the endpoints:
    \begin{equation*}
    x - y = [\underline{x} - \overline{y}, \overline{x} - \underline{y}]
    \end{equation*}
    
    \item[Multiplication.] The product is the interval spanning the minimum and maximum of the four products of the endpoints.
    \begin{equation*}
    x \cdot y = [\underline{p}, \overline{p}],
    \end{equation*}
    where
    \begin{align*}
        \underline{p} &= \min(\underline{x}\,\underline{y}, \underline{x}\,\overline{y}, \overline{x}\,\underline{y}, \overline{x}\,\overline{y}), \\
        \overline{p} &= \max(\underline{x}\,\underline{y}, \underline{x}\,\overline{y}, \overline{x}\,\underline{y}, \overline{x}\,\overline{y}).
    \end{align*}
    
    \item[Division.] Division is defined as multiplication by the reciprocal of $y$, provided that $0 \notin y$.
    $$ \frac{x}{y} = x \cdot \left[\frac{1}{\overline{y}}, \frac{1}{\underline{y}}\right], \quad \text{if } 0 \notin [\underline{y}, \overline{y}] $$
    Suppose the divisor interval $y$ contains zero. In that case, the operation is typically undefined or results in extended intervals (e.g., a union of two intervals), depending on the specific framework being used \cite{moore2009introduction}.
\end{description}

\subsection{Interval Matrix Multiplication}

The operations of IA extend naturally to linear algebra, enabling computations with interval matrices. An interval matrix $\mathbf{A}$ is a matrix whose elements are intervals. Given two compatible interval matrices, $\mathbf{A}$ and $\mathbf{B}$, their product $\mathbf{C} = \mathbf{A}\mathbf{B}$ is an interval matrix whose elements $c_{ij}$ are defined by the interval extension of the standard dot product:
\begin{equation*}
c_{ij} = \sum_{k=1}^{n} a_{ik} b_{kj}.
\end{equation*}
Here, each multiplication $a_{ik} b_{kj}$ is an interval multiplication, and the summation $\sum$ represents a sequence of interval additions. This operation is fundamental for solving interval linear systems, which are crucial for analyzing the effects of bounded uncertainties in models. However, this definition is highly susceptible to the dependency problem; if the same interval variable appears in multiple entries of $\mathbf{A}$ or $\mathbf{B}$, the resulting bounds on $\mathbf{C}$ can be much wider than the true range.

\subsection{Interval Convolutions}
IA can be extended to operations like convolutions, which are fundamental in areas such as signal processing and image analysis. By representing an input signal $\mathbf{X}$ and a filter kernel $\mathbf{W}$ as intervals, one can analyze systems where inputs or parameters are subject to bounded uncertainties.

A key operation is the interval convolution. For a 2D input $\mathbf{X}$, an interval kernel (filter) $\mathbf{W}$, and an optional interval bias $\mathbf{b}$, the resulting output feature map $\mathbf{Y}$ is computed as:
\begin{equation*}
\mathbf{Y}_{ij} = \left( \sum_{k,l} \mathbf{X}_{i+k, j+l} \cdot \mathbf{W}_{k,l} \right) + \mathbf{b}.
\end{equation*}
This computation propagates the interval bounds through the filtering operation, producing a rigorous enclosure for the true output. This is particularly useful in robust state estimation and control, where system parameters or measurements are known to lie within certain bounds \cite{Jaulin2001}. In \our{}, only the layer inputs are intervals, whereas all weights (e.g., $\mathbf{W}$ and $\Delta \mathbf{W}$) are real-valued.

\subsection{Properties and Practical Considerations}

While IA provides a mechanism for bounding activations, two characteristics are notable for implementation. First, theoretically, IA requires \emph{directed rounding}, expanding the bounds slightly outward, to strictly guarantee that no precision errors are excluded. However, in the context of neural networks, these microscopic floating-point errors are negligible compared to the inherent high variance and noise of the feature activations.

Second, IA is subject to the \emph{dependency problem}. When a variable appears multiple times in an expression, each occurrence is treated independently, leading to an overestimation of the true range. For example, computing $x - x$ for $x \in [0, 1]$ yields $[-1, 1]$ rather than the true set $\{0\}$. A related source of overestimation is the \emph{wrapping effect}: when intervals are propagated through successive layers, the image of a box is repeatedly enclosed in a larger axis-aligned box. \our{} avoids both effects, as it bounds the drift of one affine layer at a time, in which each coordinate of the input enters every output only once; the resulting drift bounds are therefore exact (Appendix~\ref{appendix:derivation_internal_representation_drift}).

\section{Metrics}\label{appendix:sec_metrics}

We assess continual learning performance using AA. Let $T$ be the total number of tasks, and $R_{i,j}$ denote the test accuracy on task $i$ after the model has finished training on task $j$. Average Accuracy (AA) reports the mean performance across all tasks after the final training session:
\begin{equation*}
    AA = \frac{1}{T} \sum_{i=1}^{T} R_{i,T}.
\end{equation*}


\section{Detailed Derivation of Internal Representation Drift Loss}
\label{appendix:derivation_internal_representation_drift}

In this section, we provide the step-by-step derivation of the regularization term $\mathcal{L}_{\mathrm{IntDrift}}$. Our goal is to ensure that the parameter updates for a layer do not alter the layer's output for any input belonging to the \textit{established activation region} of previous tasks.

\paragraph{The Stability Condition.}
Consider the $l$-th affine layer of the network with weights $\mathbf{W}_l$ and bias $\mathbf{b}_l$. Let $\mathbf{z} \in \mathbb{R}^{D_{in}}$ denote the input to this layer (i.e., the output of layer $l-1$). The preactivation output is given by:
\begin{equation*}
    f(\mathbf{z}; \boldsymbol{\theta}_l) = \mathbf{W}_l \mathbf{z} + \mathbf{b}_l.
\end{equation*}
After training on a new task, the parameters update to $\mathbf{W}_l + \Delta \mathbf{W}_l$ and $\mathbf{b}_l + \Delta \mathbf{b}_l$. The drift in the preactivation for a given input $\mathbf{z}$ is:
\begin{equation*}
    \Delta f(\mathbf{z}) = f(\mathbf{z}; \boldsymbol{\theta}_l + \Delta \boldsymbol{\theta}_l) - f(\mathbf{z}; \boldsymbol{\theta}_l) = \Delta \mathbf{W}_l \mathbf{z} + \Delta \mathbf{b}_l.
\end{equation*}
To prevent catastrophic forgetting, we require this drift to be zero for all inputs residing within the hypercube $\mathcal{H}$ defined by the previous tasks. Let this hypercube be denoted by the interval vector $[\underline{\mathbf{z}}, \overline{\mathbf{z}}]$. The condition is:
\begin{equation*}
    \Delta \mathbf{W}_l \mathbf{z} + \Delta \mathbf{b}_l = \mathbf{0}, \quad \forall \mathbf{z} \in [\underline{\mathbf{z}}, \overline{\mathbf{z}}].
\end{equation*}

\paragraph{Interval Extension.}
Since checking every point $\mathbf{z}$ is impossible, we use IA to bound the drift $\Delta f(\mathbf{z})$. The interval extension of the linear map $\Delta \mathbf{W}_l \mathbf{z}$ computes the tightest box containing the image of the input hypercube.
For the $i$-th output coordinate, the drift bounds $[\underline{\delta}_i, \overline{\delta}_i]$ are computed as:
\begin{equation*}
    [\underline{\delta}_i, \overline{\delta}_i] = \Delta \mathbf{W}_{l,i} \cdot [\underline{\mathbf{z}}, \overline{\mathbf{z}}] + \Delta b_{l,i},
\end{equation*}
where $\Delta \mathbf{W}_{l,i}$ is the $i$-th row of the update matrix.

\paragraph{Decomposition into Positive and Negative Parts.}
In IA, the product of a scalar $w$ and an interval $[a, b]$ depends on the sign of $w$:
\begin{equation*}
    w \cdot [a, b] = \begin{cases} [w a, w b] & \text{if } w \ge 0 \\ [w b, w a] & \text{if } w < 0 \end{cases}.
\end{equation*}
To vectorize this operation, we decompose the weight row $\Delta \mathbf{W}_{l,i}$ into its positive and negative components:
\begin{equation*}
    \Delta \mathbf{W}_{l,i}^+ = \max(0, \Delta \mathbf{W}_{l,i}), \quad \Delta \mathbf{W}_{l,i}^- = \max(0, -\Delta \mathbf{W}_{l,i}).
\end{equation*}
Thus, $\Delta \mathbf{W}_{l,i} = \Delta \mathbf{W}_{l,i}^+ - \Delta \mathbf{W}_{l,i}^-$. Applying interval arithmetic, we obtain:
\begin{align*}
    \Delta \mathbf{W}_{l,i} \cdot [\underline{\mathbf{z}}, \overline{\mathbf{z}}]
    &= (\Delta \mathbf{W}_{l,i}^+ - \Delta \mathbf{W}_{l,i}^-) \cdot [\underline{\mathbf{z}}, \overline{\mathbf{z}}] \\
    &= [\Delta \mathbf{W}_{l,i}^+ \underline{\mathbf{z}}, \Delta \mathbf{W}_{l,i}^+ \overline{\mathbf{z}}]
    - [\Delta \mathbf{W}_{l,i}^- \underline{\mathbf{z}}, \Delta \mathbf{W}_{l,i}^- \overline{\mathbf{z}}] \\
    &= [\Delta \mathbf{W}_{l,i}^+ \underline{\mathbf{z}} - \Delta \mathbf{W}_{l,i}^- \overline{\mathbf{z}},
    \Delta \mathbf{W}_{l,i}^+ \overline{\mathbf{z}} - \Delta \mathbf{W}_{l,i}^- \underline{\mathbf{z}}].
\end{align*}
The second equality is exact because $\Delta \mathbf{W}_{l,i}^+$ and $\Delta \mathbf{W}_{l,i}^-$ have disjoint supports, so each coordinate of $\mathbf{z}$ enters only one of the two terms. Using the interval subtraction rule $[a, b] - [c, d] = [a - d, b - c]$, we derive the bounds for the drift:
\begin{align*}
    \underline{\delta}_i &= \Delta \mathbf{W}_{l,i}^+ \underline{\mathbf{z}} - \Delta \mathbf{W}_{l,i}^- \overline{\mathbf{z}} + \Delta b_{l,i}, \\
    \overline{\delta}_i &= \Delta \mathbf{W}_{l,i}^+ \overline{\mathbf{z}} - \Delta \mathbf{W}_{l,i}^- \underline{\mathbf{z}} + \Delta b_{l,i}.
\end{align*}

\paragraph{Loss Formulation.}

The condition $\Delta f(\mathbf{z}) = 0$ requires the interval $[\underline{\delta}_i, \overline{\delta}_i]$ to collapse to the singleton set $[0,0]$. To enforce this behavior, we minimize the squared magnitude of the interval bounds across all protected layers, denoted by $\mathcal{S}$:
\begin{equation*}
    \mathcal{L}_{\mathrm{IntDrift}}
    =
    \sum_{l \in \mathcal{S}}
    \sum_{i}
    \left(
        (\underline{\delta}_{l,i})^2
        +
        (\overline{\delta}_{l,i})^2
    \right).
\end{equation*}

Here, $\underline{\delta}_{l,i}$ and $\overline{\delta}_{l,i}$ denote the bounds $\underline{\delta}_i$ and $\overline{\delta}_i$ of layer $l$. For clarity, constant normalization factors are omitted from the formulation above. Minimizing this objective progressively shrinks the admissible drift interval, thereby enforcing functional consistency between the current and previous model within the protected regions of the feature space.

\section{Internal Representation Drift for Convolutional Layers}
\label{appendix:sec_internal_representation_drift_loss_for_other_layers}

\paragraph{Convolutional Layer.}
We extend the derivation of $\mathcal{L}_{\mathrm{IntDrift}}$ to 2D convolutional layers. Let $f(\mathbf{z}; \boldsymbol{\theta}_l) = \mathrm{Conv2d}(\mathbf{z}, \mathbf{W}_l) + \mathbf{b}_l$ denote the layer operation. Since convolution is a linear operator applied to local spatial patches, it can be reformulated as an affine transformation on unfolded input vectors.

Consider a single output channel $i$. The preactivation at a specific spatial location is computed by the dot product of the flattened filter kernel $\mathbf{k}_i$ (derived from $\mathbf{W}_l$) and the corresponding flattened input patch $\mathbf{z}_p$. Let $\Delta \mathbf{k}_i$ and $\Delta b_i$ represent the parameter updates. The stability condition requires the drift to be zero for any patch $\mathbf{z}_p$ within the established bounds $[\underline{\mathbf{z}}_p, \overline{\mathbf{z}}_p]$:
\begin{equation*}
    \Delta \mathbf{k}_i^\top \mathbf{z}_p + \Delta b_i = 0, \quad \forall \mathbf{z}_p \in [\underline{\mathbf{z}}_p, \overline{\mathbf{z}}_p].
\end{equation*}

By treating the flattened patch $\mathbf{z}_p$ as the input vector, this formulation becomes mathematically identical to the affine case derived in Appendix~\ref{appendix:derivation_internal_representation_drift}. We apply the same IA decomposition. Let $\Delta \mathbf{k}_i^+$ and $\Delta \mathbf{k}_i^-$ denote the positive and negative components of the filter update. The bounds on the drift $[\underline{\delta}_i, \overline{\delta}_i]$ for this channel are:
\begin{align*}
    \underline{\delta}_i &= (\Delta \mathbf{k}_i^+)^\top \underline{\mathbf{z}}_p - (\Delta \mathbf{k}_i^-)^\top \overline{\mathbf{z}}_p + \Delta b_i, \\
    \overline{\delta}_i &= (\Delta \mathbf{k}_i^+)^\top \overline{\mathbf{z}}_p - (\Delta \mathbf{k}_i^-)^\top \underline{\mathbf{z}}_p + \Delta b_i.
\end{align*}
The loss $\mathcal{L}_{\mathrm{IntDrift}}$ is then computed by minimizing the squared $L_2$ norm of these bounds, summed across all channels and spatial locations. This ensures that the convolutional filters remain functionally invariant within the protected local geometric regions of past tasks.

\paragraph{Batch Normalization Layer.}
During inference, a Batch Normalization (BN) layer operates as a channel-wise affine transformation. Let $\boldsymbol{\gamma}, \boldsymbol{\beta}$ be the learnable parameters and $\boldsymbol{\mu}, \boldsymbol{\sigma}^2$ be the frozen running statistics. The layer operation is:
\begin{equation*}
    f(\mathbf{z}; \boldsymbol{\theta}) = \underbrace{\frac{\boldsymbol{\gamma}}{\sqrt{\boldsymbol{\sigma}^2 + \epsilon}}}_{\mathbf{W}_{\mathrm{eff}}} \mathbf{z} + \underbrace{\left(\boldsymbol{\beta} - \frac{\boldsymbol{\gamma} \boldsymbol{\mu}}{\sqrt{\boldsymbol{\sigma}^2 + \epsilon}}\right)}_{\mathbf{b}_{\mathrm{eff}}}.
\end{equation*}
When the parameters update by $(\Delta \boldsymbol{\gamma}, \Delta \boldsymbol{\beta})$, the resulting functional drift is equivalent to an affine update with effective parameters:
\begin{equation*}
    \Delta \mathbf{W}_{\mathrm{eff}} = \frac{\Delta \boldsymbol{\gamma}}{\sqrt{\boldsymbol{\sigma}^2 + \epsilon}}, \quad \Delta \mathbf{b}_{\mathrm{eff}} = \Delta \boldsymbol{\beta} - \frac{\Delta \boldsymbol{\gamma} \boldsymbol{\mu}}{\sqrt{\boldsymbol{\sigma}^2 + \epsilon}}.
\end{equation*}
Since the drift form $\Delta \mathbf{W}_{\mathrm{eff}} \mathbf{z} + \Delta \mathbf{b}_{\mathrm{eff}}$ is identical to the standard affine case, we compute $\mathcal{L}_{\mathrm{IntDrift}}$ directly using the interval bounds derived in Appendix~\ref{appendix:derivation_internal_representation_drift}, substituting $\Delta \mathbf{W}_l$ and $\Delta \mathbf{b}_l$ with $\Delta \mathbf{W}_{\mathrm{eff}}$ and $\Delta \mathbf{b}_{\mathrm{eff}}$.

\section{Facts and Proofs}
\label{appendix:sec_proofs}

In this section, we provide the full mathematical formalization and proofs supporting Theorem~\ref{thm:math_guarantee} from the main text.

\paragraph{Assumptions.}
Let $f,f':\mathcal{Z}\to\mathcal{Y}$ be linear mappings from the latent representation space $\mathcal{Z}\subseteq\mathbb{R}^d$ to the output space $\mathcal{Y}\subseteq\mathbb{R}^N$, where $d$ and $N$ denote the respective dimensions. Thus, for all $\mathbf{x},\mathbf{y}\in\mathcal{Z}$ and $a,b\in\mathbb{R}$, they satisfy $f(a\mathbf{x}+b\mathbf{y})=af(\mathbf{x})+bf(\mathbf{y})$, and analogously for $f'$. We define the functional perturbation as $\Delta f\triangleq f'-f$.

Although stated for linear mappings, all results below also extend to affine mappings of the form $f(\mathbf{z})=\mathbf{W}\mathbf{z}+\mathbf{b}$ and $f'(\mathbf{z})=\mathbf{W}'\mathbf{z}+\mathbf{b}'$. In this case, $\Delta f_i(\mathbf{e}_j)$ is interpreted as the corresponding entry $\Delta W_{ij}$ of $\Delta\mathbf{W}=\mathbf{W}'-\mathbf{W}$. This extension follows because Fact~1 does not require linearity, while the bias perturbation cancels in the differences $\overline{\delta}_i-\underline{\delta}_i$ and $\Delta f_i(\mathbf{z}^*)-\Delta f_i(\tilde{\mathbf{z}})$ used in the proofs of Facts~2 and~3.

Let
$\mathcal{H} = \prod_{j=1}^d [\underline{z}_j, \overline{z}_j] \subseteq \mathcal{Z}$
denote the protected hypercube, where $\underline{z}_j$ and $\overline{z}_j$ are the lower and upper bounds of historical activations along the $j$-th coordinate, respectively. Importantly, $\mathcal{H}$ need not contain every historical representation. We define its diameter along dimension $j$ as
$\mathrm{diam}_j(\mathcal{H}) \triangleq \overline{z}_j - \underline{z}_j$
and assume $\mathrm{diam}_j(\mathcal{H}) > 0$ for all $j \in \{1,\dots,d\}$.

We further consider the internal drift loss:
\begin{equation*}
\mathcal{L}_{\mathrm{IntDrift}} \triangleq \frac{\lambda}{N} \sum_{i=1}^N \left[
\left(\sup_{\mathbf{z} \in \mathcal{H}} \Delta f_i(\mathbf{z})\right)^2
+
\left(\inf_{\mathbf{z} \in \mathcal{H}} \Delta f_i(\mathbf{z})\right)^2
\right],
\end{equation*}
where $\lambda > 0$ is the regularization weight (corresponding to $\lambda_{\mathrm{IntDrift}}$ in Eq.~\eqref{eq:reg_int_repr}), and $\Delta f_i(\mathbf{z})$ denotes the $i$-th component of the output perturbation $\Delta f(\mathbf{z})$. We assume that, for some $\varepsilon > 0$, this loss satisfies
$\mathcal{L}_{\mathrm{IntDrift}} \leq \varepsilon$.

\paragraph{Formal Statements.}
Under the above assumptions, we establish three bounds on the functional perturbation:

\begin{itemize}
    \item \textbf{Fact 1 (Drift within the protected region).}
    For any $\mathbf{z} \in \mathcal{H}$, the absolute functional perturbation is uniformly bounded:
    \begin{equation*}
        \|\Delta f(\mathbf{z})\|_\infty
        \leq \sqrt{\frac{N\varepsilon}{\lambda}}
        \triangleq \kappa.
    \end{equation*}

    \item \textbf{Fact 2 (Sensitivity along coordinate directions).}
    For any output dimension $i$ and coordinate direction $\mathbf{e}_j$, the functional perturbation satisfies:
    \begin{equation*}
        |\Delta f_i(\mathbf{e}_j)|
        \leq \frac{1}{\mathrm{diam}_j(\mathcal{H})}
        \sqrt{\frac{2N\varepsilon}{\lambda}}.
    \end{equation*}

    \item \textbf{Fact 3 (Drift outside the protected region).}
    For any $\mathbf{z}^* \notin \mathcal{H}$, define its coordinate-wise distance from the hypercube as
    \begin{equation*}
        d_j(\mathbf{z}^*) \triangleq
        \min_{\mathbf{z}' \in \mathcal{H}}
        |z_j^* - z_j'|.
    \end{equation*}
    The functional perturbation is then bounded by:
    \begin{equation*}
        |\Delta f_i(\mathbf{z}^*)|
        \leq \kappa +
        \sum_{j=1}^d
        \frac{d_j(\mathbf{z}^*)}{\mathrm{diam}_j(\mathcal{H})}
        \sqrt{\frac{2N\varepsilon}{\lambda}}.
    \end{equation*}
\end{itemize}

\paragraph{Proof of Fact 1.}
For each output dimension $i$, let
\begin{align*}
    \overline{\delta}_i &\triangleq \sup_{\mathbf{z}\in\mathcal{H}} \Delta f_i(\mathbf{z}),\\
    \underline{\delta}_i &\triangleq \inf_{\mathbf{z}\in\mathcal{H}} \Delta f_i(\mathbf{z}).
\end{align*}
Since $\mathcal{H}$ is compact and $\Delta f_i$ is continuous, both extrema are finite and attained. Thus, for any $\mathbf{z}\in\mathcal{H}$,
\begin{equation*}
    \underline{\delta}_i \leq \Delta f_i(\mathbf{z})
    \leq \overline{\delta}_i,
\end{equation*}
which implies
\begin{equation*}
    |\Delta f_i(\mathbf{z})|^2
    \leq \max\left\{
    (\underline{\delta}_i)^2,(\overline{\delta}_i)^2
    \right\}
    \leq (\underline{\delta}_i)^2+(\overline{\delta}_i)^2.
\end{equation*}

By the assumption $\mathcal{L}_{\mathrm{IntDrift}}\leq\varepsilon$,
\begin{equation*}
    \frac{\lambda}{N}\sum_{k=1}^N
    \left[
    (\overline{\delta}_k)^2+(\underline{\delta}_k)^2
    \right]
    \leq \varepsilon.
\end{equation*}
Since all terms are non-negative, each individual term satisfies
\begin{equation*}
    (\overline{\delta}_i)^2+(\underline{\delta}_i)^2
    \leq \frac{N\varepsilon}{\lambda}.
\end{equation*}
Combining the above inequalities, we obtain, for every $\mathbf{z}\in\mathcal{H}$ and output dimension $i$,
\begin{equation*}
    |\Delta f_i(\mathbf{z})|
    \leq \sqrt{\frac{N\varepsilon}{\lambda}}.
\end{equation*}
Taking the maximum over all output dimensions yields
\begin{equation*}
    \|\Delta f(\mathbf{z})\|_\infty
    \leq \sqrt{\frac{N\varepsilon}{\lambda}}
    \triangleq \kappa,
\end{equation*}
which completes the proof. \qed

\paragraph{Proof of Fact 2.}
By linearity, any $\mathbf{z}\in\mathcal{Z}$ can be expressed in the canonical basis as $\mathbf{z}=\sum_{j=1}^d z_j\mathbf{e}_j$, giving
\begin{equation*}
    \Delta f_i(\mathbf{z})
    = \sum_{j=1}^d z_j\Delta f_i(\mathbf{e}_j).
\end{equation*}
Since $\mathcal{H}$ is a Cartesian product of independent coordinate intervals, the supremum and infimum of this linear expression decompose across dimensions:
\begin{align*}
    \overline{\delta}_i
    &= \sup_{\mathbf{z}\in\mathcal{H}} \Delta f_i(\mathbf{z})
    = \sum_{j=1}^d \max_{z_j\in[\underline{z}_j,\overline{z}_j]}
    z_j\Delta f_i(\mathbf{e}_j),\\
    \underline{\delta}_i
    &= \inf_{\mathbf{z}\in\mathcal{H}} \Delta f_i(\mathbf{z})
    = \sum_{j=1}^d \min_{z_j\in[\underline{z}_j,\overline{z}_j]}
    z_j\Delta f_i(\mathbf{e}_j).
\end{align*}
Using the identity $\max_{x\in[a,b]} cx-\min_{x\in[a,b]} cx=|c|(b-a)$, we obtain
\begin{equation*}
    \overline{\delta}_i-\underline{\delta}_i
    = \sum_{j=1}^d |\Delta f_i(\mathbf{e}_j)|
    \mathrm{diam}_j(\mathcal{H}).
\end{equation*}
Next, applying $a^2+b^2\geq\frac{1}{2}(a-b)^2$ yields
\begin{align*}
    (\overline{\delta}_i)^2+(\underline{\delta}_i)^2
    &\geq \frac{1}{2}(\overline{\delta}_i-\underline{\delta}_i)^2\\
    &= \frac{1}{2}\left(
    \sum_{j=1}^d |\Delta f_i(\mathbf{e}_j)|
    \mathrm{diam}_j(\mathcal{H})
    \right)^2.
\end{align*}
By the assumption $\mathcal{L}_{\mathrm{IntDrift}}\leq\varepsilon$ and the non-negativity of all terms in its definition, each output dimension satisfies
\begin{equation*}
    (\overline{\delta}_i)^2+(\underline{\delta}_i)^2
    \leq \frac{N\varepsilon}{\lambda}.
\end{equation*}
Combining the above inequalities gives
\begin{equation*}
    \frac{1}{2}\left(
    \sum_{j=1}^d |\Delta f_i(\mathbf{e}_j)|
    \mathrm{diam}_j(\mathcal{H})
    \right)^2
    \leq \frac{N\varepsilon}{\lambda}.
\end{equation*}
Since every summand is non-negative, the sum dominates each individual term. Therefore, for every $j$,
\begin{equation*}
    \frac{1}{2}
    \left(
    |\Delta f_i(\mathbf{e}_j)|
    \mathrm{diam}_j(\mathcal{H})
    \right)^2
    \leq \frac{N\varepsilon}{\lambda}.
\end{equation*}
Finally, using $\mathrm{diam}_j(\mathcal{H})>0$ and taking square roots, we obtain
\begin{equation*}
    |\Delta f_i(\mathbf{e}_j)|
    \leq \frac{1}{\mathrm{diam}_j(\mathcal{H})}
    \sqrt{\frac{2N\varepsilon}{\lambda}},
\end{equation*}
which completes the proof. \qed

\paragraph{Proof of Fact 3.}
To demonstrate that this directional bound preserves functional plasticity for novel tasks, we evaluate the permissible functional drift for a $\mathbf{z}^* \notin \mathcal{H}$. 

Let $\mathbf{z}^*$ possess features in the complement of the historical intervals. We define its distance to hypercube $\mathcal{H}$ along the $j$-th dimension as $d_j(\mathbf{z}^*) \triangleq \min_{\mathbf{z}' \in \mathcal{H}} |z^*_j - z'_j|$. 

Let $\tilde{\mathbf{z}} \in \mathcal{H}$ be the closest boundary point to $\mathbf{z}^*$, i.e., $\tilde{z}_j = \min(\max(z^*_j, \underline{z}_j), \overline{z}_j)$ for all $j$. By the properties of vector spaces, $\mathbf{z}^* = \tilde{\mathbf{z}} + (\mathbf{z}^* - \tilde{\mathbf{z}})$. Due to the linearity of the operator, we can perfectly decompose the functional drift:
\begin{equation*}
\Delta f_i(\mathbf{z}^*) = \Delta f_i(\tilde{\mathbf{z}}) + \Delta f_i(\mathbf{z}^* - \tilde{\mathbf{z}}) = \Delta f_i(\tilde{\mathbf{z}}) + \sum_{j=1}^d \Delta f_i(\mathbf{e}_j) (z^*_j - \tilde{z}_j)
\end{equation*}
Taking the absolute magnitude and applying the triangle inequality:
\begin{equation*}
|\Delta f_i(\mathbf{z}^*)| \leq |\Delta f_i(\tilde{\mathbf{z}})| + \sum_{j=1}^d |\Delta f_i(\mathbf{e}_j)| |z^*_j - \tilde{z}_j|
\end{equation*}
By the definition of the closest point $\tilde{\mathbf{z}}$, the term $|z^*_j - \tilde{z}_j|$ is exactly the minimum distance $d_j(\mathbf{z}^*)$. From Fact 1, since $\tilde{\mathbf{z}} \in \mathcal{H}$, we know that $|\Delta f_i(\tilde{\mathbf{z}})| \leq \kappa$. Substituting our bound from Fact 2 yields the maximum permissible functional plasticity for the novel representation:
\begin{equation*}
|\Delta f_i(\mathbf{z}^*)| \leq \kappa + \sum_{j=1}^d \left( \frac{d_j(\mathbf{z}^*)}{\mathrm{diam}_j(\mathcal{H})} \right) \sqrt{\frac{2 N \varepsilon}{\lambda}}
\end{equation*} \qed

\subsection{Proof of Theorem~\ref{thm:math_guarantee}}
\paragraph{Proof.}
The theorem follows directly from Facts~1 and~3. Consider an arbitrary representation $\mathbf{z}\in\mathcal{Z}$.

\textit{Case 1: $\mathbf{z}\in\mathcal{H}$.}
In this case, $d_j(\mathbf{z})=0$ for all $j$. By Fact~1,
\begin{equation*}
    |\Delta f_i(\mathbf{z})|
    \leq \kappa
    = \sqrt{\frac{N\varepsilon}{\lambda}}
    = \mathcal{O}(\sqrt{\varepsilon}),
\end{equation*}
which recovers the theorem's bound, as the distance-dependent term vanishes.

\textit{Case 2: $\mathbf{z}\notin\mathcal{H}$.}
Applying Fact~3 directly gives
\begin{equation*}
    |\Delta f_i(\mathbf{z})|
    \leq \sqrt{\frac{N\varepsilon}{\lambda}}
    + \sum_{j=1}^d
    \frac{d_j(\mathbf{z})}{\mathrm{diam}_j(\mathcal{H})}
    \sqrt{\frac{2N\varepsilon}{\lambda}}.
\end{equation*}
Since $N$ and $\lambda$ are fixed, both coefficients scale as $\mathcal{O}(\sqrt{\varepsilon})$. Consequently,
\begin{equation*}
    |\Delta f_i(\mathbf{z})|
    \leq \mathcal{O}(\sqrt{\varepsilon})
    + \sum_{j=1}^d
    \frac{d_j(\mathbf{z})}{\mathrm{diam}_j(\mathcal{H})}
    \mathcal{O}(\sqrt{\varepsilon}),
\end{equation*}
which establishes the claimed bound in Theorem~\ref{thm:math_guarantee}. \qed

\subsection{Extension to Lipschitz Activation Functions}
\label{appendix}

Theorem~\ref{thm:math_guarantee} bounds the drift of the pre-activation outputs $f(\mathbf{z})$ of a protected layer. Consider an element-wise activation function $\sigma$ that is $L_\sigma$-Lipschitz, i.e., $|\sigma(a) - \sigma(b)| \leq L_\sigma |a - b|$ for all $a,b \in \mathbb{R}$. Then, for every output unit $i$,
\begin{equation*}
\big| \sigma\big(f'_i(\mathbf{z})\big) - \sigma\big(f_i(\mathbf{z})\big) \big|
\leq
L_\sigma \, |\Delta f_i(\mathbf{z})|.
\end{equation*}
Consequently, the bounds in Fact~1, Fact~3, and Theorem~\ref{thm:math_guarantee} also apply to the post-activation outputs, with their right-hand sides scaled by $L_\sigma$. For ReLU and leaky ReLU, $L_\sigma=1$, so these activations do not amplify functional drift. For GELU, $L_\sigma \approx 1.13$.

\section{Training Algorithm}
\label{appendix:sec_training_algorithm}

The complete training procedure for \our{} is detailed in Algorithm~\ref{appendix:alg_training}.

\paragraph{Optimization Setup.}
We optimize the set of learnable parameters $\boldsymbol{\Psi} = \{\boldsymbol{\phi}, \mathbf{w}\}$, comprising the prompt pool parameters $\boldsymbol{\phi}$ and the classifier weights $\mathbf{w}$, while keeping the backbone $f_{\boldsymbol{\theta}}$ frozen. The training process incorporates a standard prompt-query mechanism \textsc{QueryPrompts} that retrieves task-relevant prompts and computes any method-specific auxiliary losses (e.g., orthogonality penalties), denoted as $\mathcal{L}_{\mathrm{aux}}$.

\paragraph{Training Dynamics.}
The optimization strategy gradually balances stability and adaptability across tasks. During the first task ($t=1$), the model optimizes the classification loss together with the variance regularization $\mathcal{L}_{\mathrm{Var}}$, which is crucial for compressing early task activations into compact, well-defined regions. For subsequent tasks ($t \ge 2$), the full regularization is applied by incorporating $\mathcal{L}_{\mathrm{IntDrift}}$, $\mathcal{L}_{\mathrm{Feat}}$, and $\mathcal{L}_{\mathrm{Align}}$. These terms encourage the model to produce consistent outputs for previously seen data, ensuring that new learning does not interfere with the predictions established for earlier tasks.

\begin{algorithm}[htbp]
\caption{\our{} Training Procedure}
\label{appendix:alg_training}
\footnotesize
\begin{algorithmic}[1]
\REQUIRE Sequence of $T$ tasks $\{\mathcal{D}_t\}_{t=1}^T$, Frozen backbone $f_{\boldsymbol{\theta}}$.
\REQUIRE Learnable classification head $g(\cdot; \mathbf{w})$ and prompts $\boldsymbol{\phi}$.
\REQUIRE Learning rate $\eta$, number of epochs $E$, Hyperparameters $\{\lambda_{\mathrm{IntDrift}}, \lambda_{\mathrm{Feat}}, \lambda_{\mathrm{Var}}, \lambda_{\mathrm{Align}}, \lambda_{\mathrm{Prompt}}, \alpha\}$.
\ENSURE Optimized parameters $\boldsymbol{\Psi} = \{\boldsymbol{\phi}, \mathbf{w}\}$.

\STATE \textbf{Initialize:} $\mathcal{S}$ as the subset of layers for hypercube tracking; $\mathcal{H}^{(\le 0)}_l \leftarrow \emptyset$ for all $l \in \mathcal{S}$.

\FOR{task $t = 1$ to $T$}
    \FOR{epoch $e=1$ to $E$}
        \FOR{batch $(\mathbf{x}, y) \sim \mathcal{D}_t$}
            \STATE $\mathbf{x}_p, \mathcal{L}_{\mathrm{aux}} \leftarrow \textsc{QueryPrompts}(\mathbf{x})$ \COMMENT{Retrieve task-specific prompts}
            \STATE $\mathbf{z} \leftarrow f_{\boldsymbol{\theta}}(\mathbf{x}_p)$ \COMMENT{Representation from backbone}
            \STATE $\hat{\mathbf{y}} \leftarrow g(\mathbf{z}; \mathbf{w})$ \COMMENT{Forward pass through classification head}
            
            \STATE $\mathcal{L}_{\mathrm{Total}} \leftarrow \mathcal{L}_{\mathrm{CE}}(y, \hat{\mathbf{y}}) + \lambda_{\mathrm{Prompt}} \mathcal{L}_{\mathrm{aux}} + \mathcal{L}_{\mathrm{Var}}$
            
            \IF{$t > 1$}
                \STATE $\mathcal{L}_{\mathrm{stab}} \leftarrow \mathcal{L}_{\mathrm{Feat}} + \mathcal{L}_{\mathrm{IntDrift}} + \mathcal{L}_{\mathrm{Align}}$
                \STATE $\mathcal{L}_{\mathrm{Total}} \leftarrow \mathcal{L}_{\mathrm{Total}} + \mathcal{L}_{\mathrm{stab}}$
            \ENDIF
            
            \STATE $\boldsymbol{\Psi} \leftarrow \boldsymbol{\Psi} - \eta \nabla_{\boldsymbol{\Psi}} \mathcal{L}_{\mathrm{Total}}$
        \ENDFOR
    \ENDFOR

    \STATE \textbf{Update Cumulative Hypercube:}
    \FOR{each layer $l \in \mathcal{S}$}
        \STATE $\mathcal{H}_{t,l} \leftarrow \textsc{PercentileBounds}(\mathcal{Z}_{t,l}, \alpha)$ \COMMENT{$\mathcal{Z}_{t,l}$: representations of task $t$ at layer $l$}
        \IF{$\mathcal{H}^{(\le t-1)}_l = \emptyset$}
            \STATE $\mathcal{H}^{(\le t)}_l \leftarrow \mathcal{H}_{t,l}$
        \ELSE
            \STATE $\mathcal{H}^{(\le t)}_l \leftarrow \mathcal{H}^{(\le t-1)}_l \sqcup \mathcal{H}_{t,l}$ \COMMENT{Expand hypercubes via interval hull}
        \ENDIF
    \ENDFOR
\ENDFOR
\end{algorithmic}
\end{algorithm}

\section{\our{} in Other Prompt-based Methods}\label{appendix:sec_intact_in_other_methods}

Table~\ref{appendix:tab_intact_other_prompt_methods} reports the results of integrating \our{} into L2P, DualPrompt, and CODA-Prompt on DomainNet and ImageNet-R. We use the optimal hyperparameter settings for class-incremental learning (CIL) identified in the comprehensive CODA-Prompt study~\cite{smith2023CODAPrompt}, applying them consistently to both the baseline methods and their \our{}-augmented variants.

The results demonstrate that \our{} provides a modular enhancement to existing prompt-based frameworks, yielding substantial performance gains. On ImageNet-R, it improves the AA of L2P by $\mathbf{+8.0}$ p.p., DualPrompt by $\mathbf{+8.5}$ p.p., and CODA-Prompt by $\mathbf{+13.0}$ p.p.; on DomainNet, the gains are smaller ($+4.1$ to $+6.0$ p.p.). These improvements indicate that \our{} helps existing methods better balance stability and plasticity. Notably, the gains are larger than those obtained when integrating \our{} into KA-Prompt. A possible explanation is that KA-Prompt maintains an exponential moving average (EMA) of its classifier weights, which helps mitigate classifier drift, whereas L2P, DualPrompt, and CODA-Prompt do not include a comparable mechanism. We also use CODA-Prompt + \our{} for the ablation and sensitivity analyses in Sec.~\ref{sec:ablation} and~\ref{sec:sensitivity}. On ImageNet-C and ImageNet-Mix, this configuration achieves $\res{85.85}{0.34}$ and $\res{74.20}{0.28}$, respectively.

Finally, the results in this table are not directly comparable in absolute terms with those in Table~\ref{tab:main_results}, as the experiments use different backbone configurations. All methods in this table use ViT-B/16 weights fine-tuned on ImageNet-1k.

\begin{table}[t]
\centering
\caption{\textbf{Performance results of integrating \our{} with L2P, DualPrompt, and CODA-Prompt.} We report AA for ImageNet-R and DomainNet in the DIL setting. Incorporating \our{} consistently yields performance gains over prompt-based baselines. Results are averaged over 5 independent runs.}
\label{appendix:tab_intact_other_prompt_methods}
\begin{tabular}{lcc>{\columncolor{avgcolor}}c}
\toprule
Method & DomainNet & ImageNet-R & \textbf{Average} \\
\midrule
L2P & $\res{48.44}{0.09}$ & $\res{50.34}{0.26}$ & $\res{49.39}{0.18}$ \\
\textbf{+ \our{}} & $\mathbf{\res{53.44}{0.02}}$ & $\mathbf{\res{58.34}{0.24}}$ & $\mathbf{\res{55.89}{0.13}}$ \\
\midrule
DualPrompt & $\res{50.87}{0.31}$ & $\res{53.31}{0.36}$ & $\res{52.09}{0.34}$ \\
\textbf{+ \our{}} & $\mathbf{\res{56.83}{0.28}}$ & $\mathbf{\res{61.85}{0.45}}$ & $\mathbf{\res{59.34}{0.37}}$ \\
\midrule
CODA-Prompt & $\res{52.52}{0.22}$ & $\res{61.21}{0.49}$ & $\res{56.87}{0.36}$ \\
\textbf{+ \our{}} & $\mathbf{\res{56.66}{0.39}}$ & $\mathbf{\res{74.23}{0.16}}$ & $\mathbf{\res{65.45}{0.28}}$ \\
\bottomrule
\end{tabular}
\end{table}

\section{Additional Results}
\label{appendix:additional_results}

We provide additional results and analyses to further assess the effectiveness of \our{}.

\paragraph{Pairwise Ablation.}
Tab.~\ref{tab:pairwise_ablation} extends the ablation in Tab.~\ref{tab:ablation_results} by evaluating all six pairwise removals of loss components on ImageNet-R. The strongest interaction occurs between $\mathcal{L}_{\mathrm{IntDrift}}$ and $\mathcal{L}_{\mathrm{Var}}$: removing both reduces AA by $8.43$ p.p., nearly twice the combined drop from removing each individually ($4.43$ p.p.). This suggests that the two components reinforce each other, with $\mathcal{L}_{\mathrm{Var}}$ shaping the intervals whose functional drift is controlled by $\mathcal{L}_{\mathrm{IntDrift}}$. The other pairwise removals result in drops close to or smaller than the sum of the corresponding individual drops, suggesting that these components play complementary, but partially overlapping, roles.

\begin{table}[htbp]
    \centering
    \caption{\textbf{Pairwise ablation on ImageNet-R.} Extension of Tab.~\ref{tab:ablation_results} with the joint removal of every pair of loss components. We report the performance drop ($\Delta$) relative to the full \our{}.}
    \label{tab:pairwise_ablation}
    \begin{tabular}{ccccc>{\columncolor{avgcolor}}c}
    \toprule
    $\mathcal{L}_{\mathrm{IntDrift}}$ & $\mathcal{L}_{\mathrm{Var}}$ & $\mathcal{L}_{\mathrm{Align}}$ & $\mathcal{L}_{\mathrm{Feat}}$ & \textbf{AA (\%)} & \textbf{$\Delta$} \\
    \midrule
    \checkmark & \checkmark & \checkmark & \checkmark & \textbf{74.23} & --- \\
    \midrule
    $\times$ & \checkmark & \checkmark & \checkmark & 70.75 & -3.48 \\
    \checkmark & $\times$ & \checkmark & \checkmark & 73.28 & -0.95 \\
    \checkmark & \checkmark & $\times$ & \checkmark & 73.76 & -0.47 \\
    \checkmark & \checkmark & \checkmark & $\times$ & 72.44 & -1.79 \\
    \midrule
    \checkmark & $\times$ & $\times$ & \checkmark & 73.08 & -1.15 \\
    \checkmark & $\times$ & \checkmark & $\times$ & 72.03 & -2.20 \\
    \checkmark & \checkmark & $\times$ & $\times$ & 72.92 & -1.31 \\
    $\times$ & $\times$ & \checkmark & \checkmark & 65.80 & -8.43 \\
    $\times$ & \checkmark & $\times$ & \checkmark & 70.18 & -4.05 \\
    $\times$ & \checkmark & \checkmark & $\times$ & 70.72 & -3.51 \\
    \bottomrule
    \end{tabular}
\end{table}

\paragraph{Gradient Interaction between $\mathcal{L}_{\mathrm{IntDrift}}$ and $\mathcal{L}_{\mathrm{Align}}$.}
To examine whether the alignment objective conflicts with the interval drift constraint, we measure the cosine similarity between the gradients of $\mathcal{L}_{\mathrm{IntDrift}}$ and $\mathcal{L}_{\mathrm{Align}}$ with respect to the trainable parameters at every training iteration of CODA-Prompt + \our{} on ImageNet-R. For the first three tasks in which both objectives are active ($t=2,3,4$), the average cosine similarities are $0.007$, $0.012$, and $0.007$, respectively. Although occasional negative values occur, they remain small (minimum $-0.067$). These results indicate that the two objectives are largely orthogonal and do not systematically oppose each other. In particular, $\mathcal{L}_{\mathrm{Align}}$ acts as a mild geometric regularizer without consistently driving new representations toward regions strongly constrained by $\mathcal{L}_{\mathrm{IntDrift}}$. This interpretation is consistent with the pairwise ablation results, where removing both terms leads to a performance drop close to the sum of their individual effects (Tab.~\ref{tab:pairwise_ablation}).

\paragraph{Comparison with HINT.}
We additionally compare \our{} with HINT~\cite{2025krukowski+5}, an interval-based continual learning method. We focus on HINT rather than InterContiNet~\cite{wolczyk2022continual}, as HINT outperforms it across the benchmarks considered in~\cite{2025krukowski+5}. To ensure a controlled comparison, we implement HINT within the CODA-Prompt framework, keeping all other components unchanged. Following the original formulation, a universal embedding is passed through a hypernetwork (a single-layer MLP with 50 hidden neurons) to generate the classifier weights. A perturbation radius defines a hypercube around each task embedding. The hypercubes associated with the current and previous tasks are then intersected, and the universal embedding is set to the center of this intersection.

We grid-search the embedding dimension ($\{48, 96\}$), perturbation radius ($\{0.05, 0.1\}$), and hypernetwork regularization weight ($\{0.01, 0.1\}$). The best configurations are $(48, 0.1, 0.01)$ for ImageNet-R and $(48, 0.1, 0.1)$ for DomainNet. As reported in Tab.~\ref{tab:hint_comparison}, \our{} outperforms HINT by $32.8$ p.p. on ImageNet-R and $4.6$ p.p. on DomainNet.

\begin{table}[htbp]
    \centering
    \caption{\textbf{Comparison with HINT.} AA of HINT and \our{}, both implemented in the CODA-Prompt framework.}
    \label{tab:hint_comparison}
    \begin{tabular}{lcc}
        \toprule
        Method & ImageNet-R & DomainNet \\
        \midrule
        HINT~\cite{2025krukowski+5} & $\res{41.42}{1.26}$ & $\res{52.02}{0.08}$ \\
        \our{} & $\mathbf{\res{74.23}{0.16}}$ & $\mathbf{\res{56.66}{0.39}}$ \\
        \bottomrule
    \end{tabular}
\end{table}

\paragraph{Class-Incremental Learning.}

Tab.~\ref{tab:class_incremental} reports results on Split MNIST, Split Fashion-MNIST (Split~FMNIST), and Split CIFAR-10. To enable controlled comparisons within a reasonable computational budget, we evaluate all methods using the same two-layer MLP architecture, with 400 neurons per layer. Across all three benchmarks, \our{} consistently outperforms standard parameter-space regularization and feature-distillation baselines (EWC, LwF, MAS, and SI).

The gains are particularly pronounced on Split MNIST and Split FMNIST, where \our{} surpasses the strongest baseline, LwF, by $6.64$ and $8.33$ p.p., respectively. On the more challenging Split CIFAR-10 benchmark, \our{} also achieves the highest AA of $31.28\%$, outperforming the strongest baseline, EWC. These results demonstrate that activation-space constraints can effectively mitigate catastrophic forgetting beyond the domain-incremental setting.

\begin{table}[htbp]
    \centering
    \caption{AA in CIL settings. All results are averaged over 5 seeds.}
    \begin{tabular}{lccc}
        \toprule
        Method & Split MNIST & Split FMNIST & Split CIFAR-10 \\
        \midrule
        EWC & 20.57 & 19.92 & 30.43 \\
        LwF & 28.63 & 25.92 & 24.26 \\
        MAS & 17.91 & 18.74 & 21.75 \\
        SI  & 18.40 & 17.30 & 20.18 \\
        \our{} (Ours) & \textbf{35.27} & \textbf{34.25} & \textbf{31.28} \\
        \bottomrule
    \end{tabular}
    \label{tab:class_incremental}
\end{table}

\paragraph{Hypercube Coverage and Hyperparameter $\alpha$.}

The hyperparameter $\alpha$ controls the coverage of the cumulative hypercube $\mathcal{H}$ by determining how many extreme activations are excluded from each side of the coordinate intervals. Smaller values of $\alpha$ yield tighter bounds around the core task distribution but leave more outliers outside $\mathcal{H}$, introducing a trade-off between coverage and compactness. Importantly, Theorem~\ref{thm:math_guarantee} also bounds functional drift outside the hypercube, with the bound increasing proportionally to the distance from $\mathcal{H}$.

\begin{table}[htbp]
    \centering
    \caption{Sensitivity analysis of the hyperparameter $\alpha$ for CODA-Prompt + \our{} on ImageNet-R.}
    \begin{tabular}{lccccccc}
        \toprule
        $\alpha$ & 0.75 & 0.80 & 0.85 & 0.90 & 0.95 & 0.99 & 1.00 \\
        \midrule
        AA (\%) & 74.66 & 74.51 & 74.49 & 74.63 & 74.23 & 73.12 & 71.37 \\
        \bottomrule
    \end{tabular}
    \label{appendix:tab_alpha_sensitivity}
\end{table}

As shown in Tab.~\ref{appendix:tab_alpha_sensitivity}, performance on ImageNet-R remains stable across $\alpha \in [0.75, 0.95]$, but declines as $\alpha$ approaches $1.00$. Including extreme outliers in the hypercube can substantially expand its bounds into sparsely occupied feature regions. This, in turn, imposes overly restrictive constraints on the model, limiting the plasticity needed to adapt to new tasks.





\paragraph{Cumulative Hypercube Growth.}
\label{para:hypercube_growth}
To assess whether the protected region expands uncontrollably over long task sequences, we track the average coordinate diameter of the cumulative hypercube, $\frac{1}{d} \sum_{j=1}^d \mathrm{diam}_j(\mathcal{H})$, throughout the 15-task ImageNet-R benchmark using CODA-Prompt + \our{}.

\begin{table}[htbp]
    \centering
    \caption{Growth of the cumulative hypercube's average coordinate diameter across ImageNet-R.}
    \begin{tabular}{lc}
        \toprule
        Task Interval & $\frac{1}{d} \sum \mathrm{diam}_j(\mathcal{H})$ \\
        \midrule
        Tasks 1--5   & $3.280 \rightarrow 4.095$ \\
        Tasks 6--10  & $4.096 \rightarrow 4.196$ \\
        Tasks 11--15 & $4.197 \rightarrow 4.202$ \\
        \bottomrule
    \end{tabular}
    \label{tab:hypercube_growth}
\end{table}

Tab.~\ref{tab:hypercube_growth} shows that most of the expansion occurs during the first five tasks, after which the hypercube rapidly approaches saturation. Its average coordinate diameter increases by only $0.107$ over the remaining ten tasks. This indicates that the protected region stops growing as new tasks arrive, rather than continuing to expand into increasingly unoccupied feature space.

To determine whether this behavior persists over longer task sequences, we repeat the analysis on ImageNet-Mix, which comprises 30 tasks. As reported in Tab.~\ref{tab:hypercube_growth_mix}, most of the expansion again occurs early: the average coordinate diameter increases by $0.653$ over the first ten tasks, but by only $0.040$ over the subsequent twenty tasks.

\begin{table}[htbp]
    \centering
    \caption{Growth of the cumulative hypercube's average coordinate diameter across ImageNet-Mix (30 tasks), reported as mean $\pm$ standard deviation over runs.}
    \begin{tabular}{lc}
        \toprule
        Task Interval & $\frac{1}{d} \sum \mathrm{diam}_j(\mathcal{H})$ \\
        \midrule
        Tasks 1--10  & $2.991 \pm 0.005 \rightarrow 3.644 \pm 0.037$ \\
        Tasks 11--20 & $3.644 \pm 0.037 \rightarrow 3.662 \pm 0.035$ \\
        Tasks 21--30 & $3.664 \pm 0.035 \rightarrow 3.684 \pm 0.035$ \\
        \bottomrule
    \end{tabular}
    \label{tab:hypercube_growth_mix}
\end{table}

\paragraph{Occupancy and Out-of-Bounds Rate.}
To assess whether the cumulative hypercube becomes excessively loose in the high-dimensional activation space, we compare it with the task-specific hypercubes $\mathcal{H}_t$ defined in Sec.~\ref{sec:activation_interval}. We conduct this analysis on ImageNet-R using KA-Prompt with \our{}. We define occupancy as the ratio between the volume of the union of task-specific hypercubes and that of the cumulative hypercube $\mathcal{H}^{(\leq T)}$, whose bounds are given by the coordinate-wise minima and maxima across tasks. Since volume ratios scale exponentially with the dimensionality $d$, we report the dimension-normalized occupancy:
\begin{equation*}
    R_{\mathrm{norm}}
    =
    \left(
    \frac{\mathrm{Vol}\big(\bigcup_{t=1}^{T} \mathcal{H}_t\big)}{\mathrm{Vol}\big(\mathcal{H}^{(\leq T)}\big)}
    \right)^{1/d},
    \qquad
    \mathrm{Vol}(\mathcal{H}) = \prod_{j=1}^{d} \mathrm{diam}_j(\mathcal{H}).
\end{equation*}
This normalization expresses occupancy as a side-length ratio, directly relating it to the coordinate-wise bounds used in Sec.~\ref{sec:regularization}. Since computing the exact union volume is costly, we replace $\mathrm{Vol}\big(\bigcup_{t=1}^{T}\mathcal{H}_t\big)$ with $\sum_{t=1}^{T}\mathrm{Vol}(\mathcal{H}_t)$, yielding an upper bound on $R_{\mathrm{norm}}$. Indeed, $\mathrm{Vol}\big(\bigcup_t\mathcal{H}_t\big) \geq \frac{1}{T}\sum_t\mathrm{Vol}(\mathcal{H}_t)$, so the bound exceeds the exact value by at most a factor of $T^{1/d}$. For the pre-classifier activations ($d=768$, $T=15$), this factor is below $1.004$. The resulting upper bound is $88.8\%$, placing the exact occupancy between $88.5\%$ and $88.8\%$. Thus, the cumulative hypercube is only about $1.13$ times wider per dimension than the region occupied by the task-specific hypercubes.

Complementing the geometric occupancy measure, the out-of-bounds rate quantifies how often activations from new tasks fall outside the cumulative bounds established by previous tasks. For a training batch $\{\mathbf{z}_n\}_{n=1}^{N_B}$ from task $t$, evaluated at the protected layer, we define
\begin{equation*}
    R_{\mathrm{OOB}}
    =
    \frac{1}{N_B \, d}
    \sum_{n=1}^{N_B}
    \sum_{j=1}^{d}
    \mathbb{I}\Big[
        z_{n,j} < \underline{z}^{(\leq t-1)}_{j}
        \;\lor\;
        z_{n,j} > \overline{z}^{(\leq t-1)}_{j}
    \Big],
\end{equation*}
where $\underline{z}^{(\leq t-1)}_{j}$ and $\overline{z}^{(\leq t-1)}_{j}$ denote the cumulative lower and upper bounds before observing task $t$. We compute this rate over all training batches and average it across tasks $t=2,\dots,T$. On ImageNet-R, the resulting average is $R_{\mathrm{OOB}}=5.8\%$, meaning that $94.2\%$ of activation coordinates from new tasks fall within the existing bounds. Importantly, this is a coordinate-wise statistic: although most coordinates remain within the protected region, a typical sample still falls outside it along approximately $45$ of its $768$ coordinates. Thus, new tasks largely reuse existing per-coordinate ranges, limiting cumulative hypercube expansion, while individual representations can still extend beyond the protected region in a small number of dimensions.

\paragraph{Coverage of Previous-Task Activations.}
To assess how well the trained model preserves the coverage assumed in Theorem~\ref{thm:math_guarantee}, we measure the proportion of previous-task activations contained within the cumulative hypercube on ImageNet-R using CODA-Prompt + \our{}. After training task $t$, we pass the evaluation data of all tasks $s \leq t$ through the current model and compute the percentage of activation coordinates $z_j$ lying within the corresponding intervals $[\underline{z}_j, \overline{z}_j]$ of $\mathcal{H}^{(\leq t)}$. Equivalently, these are the coordinates for which $d_j(\mathbf{z})=0$ in Theorem~\ref{thm:math_guarantee}.

Tab.~\ref{tab:hypercube_coverage} reports the resulting coverage. Immediately after learning a task, at least approximately $90\%$ of its activation coordinates lie within the cumulative bounds, consistent with the quantile-based construction using $\alpha=0.95$. Coverage of previously learned tasks generally increases as training progresses, despite temporary fluctuations. For task~1, it rises from $90.3\%$ to $97.0\%$, while final coverage across tasks ranges from $96.1\%$ to $97.0\%$. Task~14 is the only exception, with a slight decrease from $96.7\%$ to $96.4\%$. Overall, no entry falls below $90\%$.

This improvement initially reflects the expansion of the cumulative hypercube. However, hypercube growth largely saturates after task~10: its average coordinate diameter increases by only $0.006$ between tasks~10 and~15 (Tab.~\ref{tab:hypercube_growth}). Despite this limited expansion, task~1's coverage continues to rise from $95.5\%$ to $97.0\%$, suggesting that its representations move further inside the protected region. The largest temporary decrease in coverage occurs after task~9, coinciding with the transition at which baseline performance drops sharply (Fig.~\ref{appendix:fig_imr_comp}).

Since coordinates contained within their intervals have $d_j(\mathbf{z})=0$, they contribute nothing to the distance-dependent term in Theorem~\ref{thm:math_guarantee}. Consequently, for previous-task representations, this term depends only on the remaining $3$--$10\%$ of coordinates lying outside the cumulative bounds.

\begin{table}[htbp]
    \centering
    \caption{\textbf{Coverage of previous-task activations on ImageNet-R.} Entry $(s,t)$ reports the percentage of activation coordinates of task $s$'s evaluation samples that fall within the cumulative hypercube $\mathcal{H}^{(\leq t)}$, evaluated using the model after training task $t$. Shaded entries correspond to evaluations performed immediately after learning task $s$.}
    \label{tab:hypercube_coverage}
    \footnotesize
    \setlength{\tabcolsep}{3pt}
    \adjustbox{max width=\linewidth}{
    \begin{tabular}{cccccccccccccccc}
    \toprule
    & \multicolumn{15}{c}{after training task $t$} \\
    \cmidrule(lr){2-16}
    task $s$ & 1 & 2 & 3 & 4 & 5 & 6 & 7 & 8 & 9 & 10 & 11 & 12 & 13 & 14 & 15 \\
    \midrule
    1 & \cellcolor{gray!15}90.3 & 92.3 & 93.7 & 94.4 & 95.6 & 95.9 & 95.6 & 96.5 & 94.5 & 95.5 & 96.4 & 96.4 & 96.9 & 97.0 & 97.0 \\
    2 &  & \cellcolor{gray!15}90.2 & 91.4 & 92.3 & 93.5 & 93.7 & 93.3 & 94.4 & 93.8 & 94.5 & 95.6 & 95.5 & 96.0 & 96.3 & 96.6 \\
    3 &  &  & \cellcolor{gray!15}92.8 & 94.0 & 95.4 & 95.5 & 95.2 & 96.0 & 94.4 & 95.2 & 96.4 & 96.3 & 96.8 & 97.1 & 97.0 \\
    4 &  &  &  & \cellcolor{gray!15}92.5 & 93.6 & 94.2 & 93.8 & 95.0 & 94.8 & 95.5 & 96.2 & 96.2 & 96.6 & 96.7 & 96.7 \\
    5 &  &  &  &  & \cellcolor{gray!15}94.2 & 94.6 & 94.1 & 95.5 & 94.1 & 95.3 & 96.1 & 95.7 & 96.5 & 96.7 & 96.6 \\
    6 &  &  &  &  &  & \cellcolor{gray!15}95.0 & 94.6 & 95.4 & 93.9 & 94.9 & 96.0 & 95.8 & 96.2 & 96.6 & 96.7 \\
    7 &  &  &  &  &  &  & \cellcolor{gray!15}96.0 & 96.7 & 94.6 & 95.4 & 96.3 & 95.8 & 96.7 & 96.9 & 96.8 \\
    8 &  &  &  &  &  &  &  & \cellcolor{gray!15}96.7 & 94.6 & 95.6 & 96.6 & 96.2 & 96.8 & 96.9 & 97.0 \\
    9 &  &  &  &  &  &  &  &  & \cellcolor{gray!15}93.2 & 94.2 & 95.4 & 94.8 & 95.6 & 96.0 & 96.1 \\
    10 &  &  &  &  &  &  &  &  &  & \cellcolor{gray!15}95.1 & 96.1 & 95.6 & 96.2 & 96.4 & 96.5 \\
    11 &  &  &  &  &  &  &  &  &  &  & \cellcolor{gray!15}96.0 & 95.8 & 96.6 & 96.7 & 96.6 \\
    12 &  &  &  &  &  &  &  &  &  &  &  & \cellcolor{gray!15}96.2 & 96.6 & 96.9 & 96.9 \\
    13 &  &  &  &  &  &  &  &  &  &  &  &  & \cellcolor{gray!15}96.4 & 96.9 & 96.6 \\
    14 &  &  &  &  &  &  &  &  &  &  &  &  &  & \cellcolor{gray!15}96.7 & 96.4 \\
    15 &  &  &  &  &  &  &  &  &  &  &  &  &  &  & \cellcolor{gray!15}97.0 \\
    \bottomrule
    \end{tabular}
    }
\end{table}

\section{Additional Visualization}
\label{appendix:sec_additional_visualization}

\paragraph{Internal Representation Stability.}
We analyze the stability of internal representations using the Split-MNIST benchmark in the DIL scenario. For this experiment, we apply \our{} to a standard MLP. To align with our framework, we treat the first hidden layer as the fixed feature extractor (applying $\mathcal{L}_{\mathrm{Feat}}$ to its outputs) and apply the rest of regularization terms ($\mathcal{L}_{\mathrm{IntDrift}}, \mathcal{L}_{\mathrm{Var}}, \mathcal{L}_{\mathrm{Align}}$) to the subsequent trainable layers.

\begin{figure}[htbp]
  \centering
  \includegraphics[width=\linewidth]{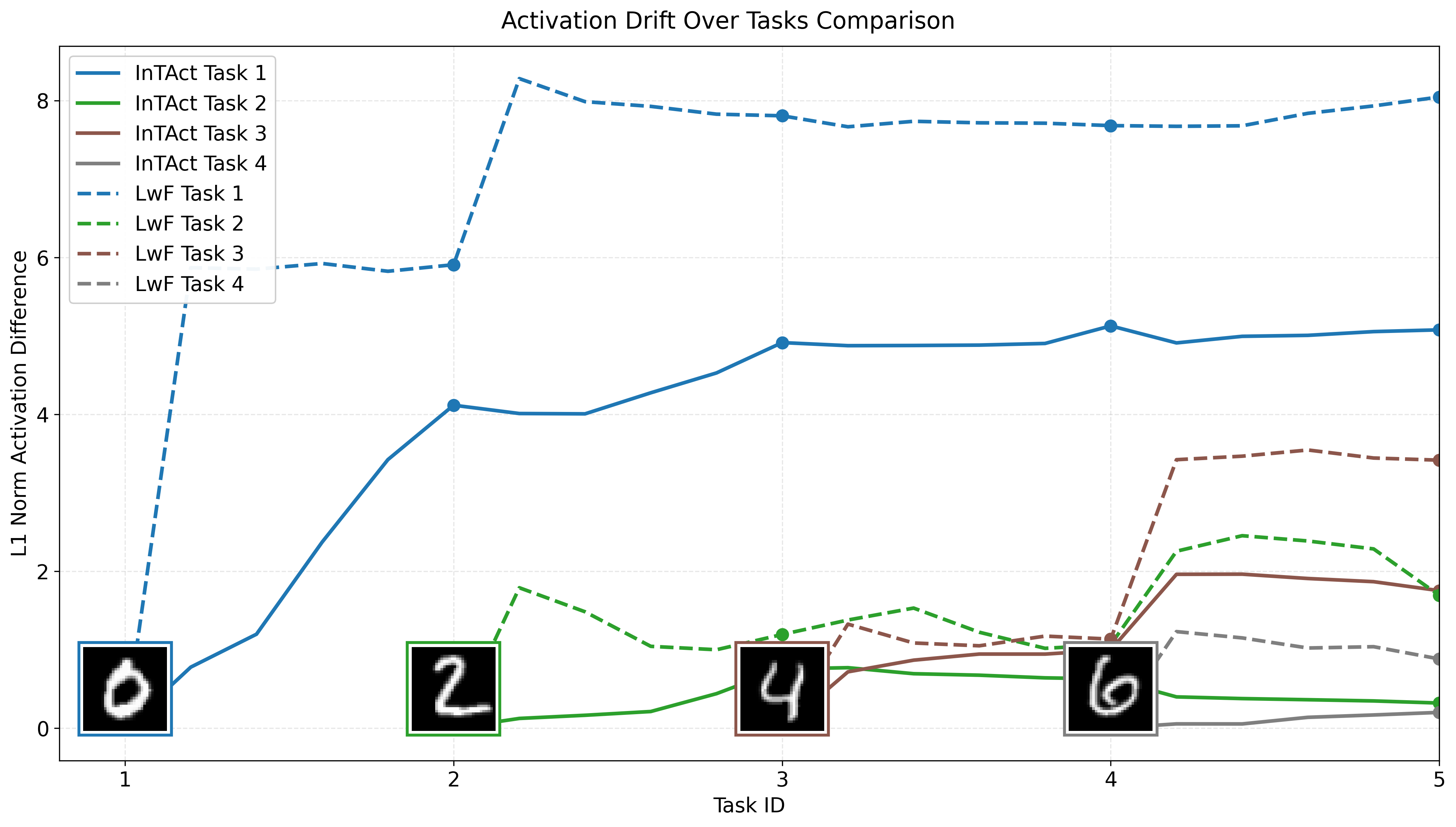}
    \caption{\textbf{Evolution of Internal Drift.} Activation stability on Split-MNIST (DIL). The plot tracks the cumulative $L_1$ shift in hidden activations for reference images from early tasks as the model trains on later tasks. \our{} successfully suppresses drift compared to LwF, validating the effectiveness of the proposed regularization.}
\label{appendix:fig_activation_drift_visualization}
\end{figure}

Fig.~\ref{appendix:fig_activation_drift_visualization} quantifies the \emph{internal representation drift} experienced by historical data as the model adapts to new tasks. We track the normalized $L_1$ difference between the activations of reference samples at the end of their initial training task and their activations after subsequent tasks. While our baseline (LwF) exhibits progressive drift, indicating that the internal features of old data are being distorted by new plasticity, \our{} effectively bounds these changes, maintaining the geometric integrity of the latent space throughout the learning sequence.

\paragraph{Conclusions.}
The comparison highlights a fundamental difference in how the two methods balance stability and plasticity. Under LwF, representational drift accumulates over the task sequence, progressively altering the features of earlier tasks and eroding previously acquired knowledge. In contrast, \our{} effectively limits this drift. Following a brief initial adaptation phase, during which the shared decision boundaries adjust to new tasks, activation shifts largely plateau. This suggests that \our{} stabilizes the network's functional behavior on previously learned tasks, mitigating the representational changes that contribute to catastrophic forgetting.

\section{Parameters and Overhead Comparison}
\label{appendix:parameters_overhead}

Table~\ref{tab:efficiency_comparison} compares computational efficiency on ImageNet-R, using reference measurements from~\cite{xu2025comp} for the baselines. For KA-Prompt and \our{} + KA-Prompt, we measure GPU memory and batch time ourselves under identical conditions: a single NVIDIA H100 (80 GB) and batch size of 128. We report peak reserved GPU memory and mean training time per batch, including data loading, on the second task, when all \our{} terms are active.

\our{} introduces \textbf{no additional learnable parameters} beyond the underlying prompt-based method while substantially improving performance. Compared with KA-Prompt, it increases peak GPU memory by only $4\%$ (from $18.87$ GB to $19.64$ GB) and training time per batch by $21\%$ (from $0.314$ s to $0.381$ s). The additional time is mainly due to $\mathcal{L}_{\mathrm{Feat}}$, which requires an extra forward pass through the frozen model. Inference remains unaffected, with nearly identical evaluation times per batch of $512$ images ($1.82$ s for KA-Prompt and $1.81$ s for \our{}). Thus, \our{} provides substantial accuracy gains at a modest training cost, without increasing inference overhead.

\begin{table*}[h]
\centering
\caption{\textbf{Efficiency comparison on ImageNet-R.} We report total parameters, trainable parameters, GPU memory usage, batch time, and AA. Baseline measurements are taken from~\cite{xu2025comp}. $^{\ddagger}$Measured by us under identical conditions on a single NVIDIA H100 (80 GB) with batch size 128. All other GPU memory and batch time measurements are from~\cite{xu2025comp} and were obtained on different hardware; therefore, the two groups are not directly comparable.}
\label{tab:efficiency_comparison}
\small
\setlength{\tabcolsep}{4pt}
\begin{tabular}{lcccc>{\columncolor{avgcolor}}c}
\toprule
Methods & Total Params & Trainable Params & GPU Memory (GB) & Batch time (s) & AA \\
\midrule
L2P & 86,263,843 & 311,387 & 19.33 & 0.66 & $\res{56.55}{0.33}$ \\
S-Prompts & 92,964,094 & 1,637,910 & 14.90 & 0.39 & $\res{27.23}{0.16}$ \\
DualPrompt & 86,514,211 & 561,755 & 19.45 & 0.66 & $\res{59.47}{1.00}$ \\
CODA-Prompt & 88,437,580 & 2,638,926 & 19.18 & 0.80 & $\res{55.21}{0.33}$ \\
CPrompt & 92,485,224 & 2,689,168 & 41.64 & 1.18 & $\res{59.48}{1.09}$ \\
C-Prompt & 89,180,505 & 3,381,851 & 11.99 & 0.42 & $\res{62.43}{0.49}$ \\
KA-Prompt & 89,180,505 & 3,381,851 & 18.87$^{\ddagger}$ & 0.31$^{\ddagger}$ & $\res{66.51}{0.36}$ \\
\midrule
\textbf{\our{} (Ours)} & 89,180,505 & 3,381,851 & 19.64$^{\ddagger}$ & 0.38$^{\ddagger}$ & $\mathbf{\res{74.23}{0.16}}$ \\
\bottomrule
\end{tabular}
\end{table*}


\newpage

\end{document}